# Variational Digital Twins

Logan A. Burnett, Umme Mahbuba Nabila, Majdi I. Radaideh*

*Department of Nuclear Engineering and Radiological Sciences, University of Michigan, Ann Arbor, MI 48109, United States*

**Abstract**

While digital twins (DT) hold promise for providing real-time insights into complex energy assets, much of the current literature either does not offer a clear framework for information exchange between the model and the asset, lacks key features needed for real-time implementation, or gives limited attention to model uncertainty. Here, we aim to solve these gaps by proposing a variational digital twin (VDT) framework that augments standard neural architectures with a single Bayesian output layer. This lightweight addition, along with a novel VDT updating algorithm, lets a twin update in seconds on commodity GPUs while producing calibrated uncertainty bounds that can inform experiment design, control algorithms, and model reliability.

The VDT is evaluated on four energy-sector problems. For critical-heat-flux prediction, uncertainty-driven active learning reaches $R^2 = 0.98$ using 47 % fewer experiments and one-third the training time of random sampling. A three-year renewable-generation twin maintains $R^2 > 0.95$ for solar output and curbs error growth for volatile wind forecasts via monthly updates that process only one month of data at a time. A nuclear reactor transient cooldown twin reconstructs thermocouple signals with $R^2 > 0.99$ and preserves accuracy after 50 % sensor loss, demonstrating robustness to degraded instrumentation. Finally, a physics-informed Li-ion battery twin, retrained after every ten discharges, lowers voltage mean-squared error by an order of magnitude relative to the best static model while adapting its credible intervals as the cell approaches end-of-life. These results demonstrate that combining modest Bayesian augmentation with efficient update schemes turns conventional surrogates into uncertainty-aware, data-efficient, and computationally tractable DTs, paving the way for dependable models across industrial and scientific energy systems.

*Keywords:* Digital Twins, Variational Inference, Power Grid Forecasting, Uncertainty Quantification, Real-time Energy Modeling

## 1. Introduction

A Digital Twin (DT) is a virtual representation of a physical asset, process, or system that is updated with real-time data to mirror the state and behavior of its physical counterpart [1]. This dynamic digital model enables simulation, analysis, and control, facilitating improved decision-making [2] and performance optimization throughout the asset's lifecycle [3, 4]. The concept has evolved significantly since its inception, becoming a cornerstone in the advancement of cyber-physical systems [5].

---

*Corresponding Authors: Logan Burnett (nucleai@umich.edu), Majdi I. Radaideh (radaideh@umich.edu)

DTs have been successfully implemented across various domains. In manufacturing [6] and complex systems [7, 8], they enable real-time monitoring, fault detection, and predictive maintenance, enhancing operational efficiency. In healthcare, DTs facilitate personalized medicine by creating patient-specific models for diagnosis and treatment planning [9]. Urban planning benefits from DTs through the creation of smart city models that assist in infrastructure management and development [10]. Despite their diverse applications, the integration of DTs in the energy sector presents unique opportunities and challenges. The remainder of this paper will focus on exploring the role of DT in energy applications, examining their potential to revolutionize energy systems through enhanced monitoring, simulation, and optimization capabilities.

*1.1. Digital Twins in the Energy Industry*

Based on our literature review, we identified five recurring gaps in major studies within the energy field related to DT: (1) proof-of-concept or simulation-only DT, (2) mislabeling and modeling inconsistency, (3) lack of real-time and adaptive learning methods, (4) absence of mathematical models, and (5) missing bidirectional synchronization.

Despite promising designs, a significant portion of DT research remains conceptual or simulation-based, without being deployed or integrated into actual physical systems. For example, Arafet and Berlanga employed deep learning models (autoencoders) to detect anomalies based on historical data [11]. While the framework demonstrates the feasibility of using time series reconstruction errors for fault detection, the implementation remains a conceptual prototype—validated offline—rather than a deployed, cyber-physically integrated DT. A related study on anomaly detection in particle accelerators utilized historical offline operational data to train autoencoders; however, the system lacked any feedback mechanism during deployment [12]. Similar patterns appear in applications such as autonomous aerial monitoring of photovoltaic (PV) plants [13], offshore wind energy frameworks [14], and nuclear plant robotics [15], where DTs are developed in virtual environments without live sensor data or control capabilities. In the context of smart cities, Li and Tan propose a simulation-based optimization framework for integrating solar energy but omit real-time validation and bi-directional communication [16]. Likewise, models for agrophotovoltaic design and geothermal heating optimization remain confined to offline simulations, lacking synchronization or feedback from live infrastructure [17, 18]. Although these studies provide valuable conceptual insights, they do not exhibit the cyber-physical integration required of a true digital twin.

A recurring issue in DT research is the misapplication of the term "digital twin" to models that do not meet the core criteria described later in Section 2, which are the typical definition or intended use of DTs. This mislabeling contributes to definitional ambiguity across the field and hinders meaningful progress toward standardization. Many studies found during this literature review labeled as DTs are better described as *digital shadows*, *high-fidelity simulators*, or *offline predictive models*. For example, the study by You and Zhu uses deep learning techniques to forecast solar market loads based on historical market data [19]. Although the authors refer to the framework as a DT, the system operates entirely offline, without sensor integration, actuation capability, and any interaction with a physical energy system. In its current form, the framework aligns more accurately with a data-driven forecasting simulator, not a DT. Similarly, Prantikos et al. present a physics-informed neural network (PINN) model to solve point kinetics equations for a nuclear reactor [20]. While the mathematical modeling is rigorous and domain-specific, the system does not interact with any live reactor, nor does it feature real-time data ingestion or bidirectional feedback. This work is best characterized as a physics-informed simulation, not a DT. Arafet and Berlanga's solar farm case study also falls into this category [11]. Although the model uses autoencoders to detect anomalies in solar farm operations, it is trained on offline historical data and lacks real-time synchronization or adaptive control capabilities. Despite being labeled a DT, the system operates as a static digital shadow.

A defining feature of a true DT is its ability to evolve over time through continuous learning from live data. However, many current DT frameworks in the energy domain lack mechanisms for real-time continuous adaptation, rendering them static in nature. This limitation is especially problematic in critical rapidly changing environments, where predictive accuracy and control effectiveness can degrade without ongoing model refinement. For example, the DT proposed by Fahim et al. for wind turbine performance forecasting leverages temporal convolutional networks (TCN) and k-nearest neighbors (kNN) regression to make real-time predictions based on Supervisory Control And Data Acquisition (SCADA) data [21]. While the system demonstrates strong predictive performance, it lacks any mechanism for model updating or re-training, making it incapable of responding to evolving turbine behavior or environmental changes. Similarly, Pimenta et al. develop a DT of an onshore wind turbine based on physics-informed simulations and SCADA-based validation [22]. Although the model is grounded in operational data, there is no evidence of continual learning or adjustment, and the system functions primarily as a static diagnostic tool. Arafet and Berlanga's work on anomaly detection in solar farms follows the same pattern: autoencoders are trained offline using historical data, and the system operates without real-time model refinement or control feedback [11]. Despite being labeled a DT, the model does not update or improve with additional sensor input, limiting its utility in dynamic environments. Li and Tan's simulation framework for smart city solar integration also lacks adaptivity [16]. The optimization models are static, and the system does not incorporate mechanisms for learning from operational outcomes or sensor feedback. Likewise, Zohdi's agrophotovoltaic design platform optimizes solar farm layouts using simulation and machine learning but remains confined to the pre-deployment phase, with no provisions for post-installation model evolution [17]. Even in the nuclear sector, where real-time adaptability is critical, similar gaps persist. The study on DT for PWR steam generators presents use cases for predictive maintenance but lacks implementation of continuous feedback loops or adaptive control strategies [23]. The models proposed are fixed in scope and do not evolve in tandem with plant behavior or operational anomalies.

The credibility and predictive accuracy of a DT are closely tied to the integration of mathematical models that capture the physical behavior of the system and propagate uncertainties from the models and the data. To function effectively, a DT must include models that define the relationships between measured inputs and controlled outputs [24]. However, many recent DT implementations either omit mathematical modeling entirely or fail to articulate it clearly in the literature. For instance, the study by Yu et al. presents a real-time simulation framework for microgrids that incorporates data flow, control architecture, and edge computing concepts [25]. While the system architecture is comprehensive, the framework does not formulate governing equations related to core microgrid dynamics such as power flow, voltage regulation, or thermal constraints. This lack of embedded mathematical modeling reduces the framework's predictive fidelity and limits its reliability in practical applications. Although methods for uncertainty quantification and propagation are well-established in the field, they are not commonly applied to DTs, either due to their high computational demands or inference/accuracy challenges, which make them unsuitable for real-time applications. Examples of such methods include model calibration and uncertainty propagation using evolutionary and swarm algorithms [26, 27], Bayesian model averaging for calibrating computer codes [28], inverse uncertainty quantification [29], and full-scale variational inference [30].

Last but not least, another often missing characteristic of a DT is bidirectional data synchronization, i.e., the ability of the DT to not only mirror the physical system, but also influence or actuate it. Across the literature, this feature remains largely unimplemented. Arafet and Berlanga's solar DT, Pimenta et al.'s wind turbine model, and even the real-time SCADA-integrated framework by Fahim et al., all ingest data in a uni-directional fashion without the capacity to send optimized control actions back to the physical layer [11, 22, 21]. Similarly, Li and Tan's smart city DT simulation and Zohdi's agrophotovoltaic optimization framework also fail to establish this two-way communication, reinforcing their classification as digital shadows or simulators [16, 17]. Conversely, the studies by [31] and [32] implemented forward control actions—specifically, rotating control drums—to regulate power in a nuclear microreactor using deep reinforcement learning and model predictive control, respectively. However, they did not provide a clear framework for how the model synchronizes with feedback signals nor how to learn from high-fidelity simulations of these systems [33]. As energy systems grow more autonomous, this gap is particularly pressing—without control feedback, these models cannot fulfill the decision-support or autonomous regulation functions expected of true digital twins.

*1.2. Novelty*

Our review of DT technologies for energy applications reveals a clear divide: existing efforts are either highly specialized—tailored to specific use cases and difficult to generalize—or overly abstract, offering vague conceptual overviews with little practical guidance. Compounding this, the lack of publicly available details—often due to proprietary tools or confidentiality—limits the ability to evaluate or replicate existing DT models. As a result, several studies conclude by highlighting key challenges [34, 35]: (1) the need for a shared definition and understanding of DT structure and function, and (2) the identification of technical hurdles in implementing DT-enabling technologies across domains. This research addresses these gaps by proposing a concrete, generalizable mathematical and computational framework for digital twins, with clearly defined components and modeling targets. Current R&D frameworks lack this level of generality and reproducibility, making it difficult to extend DT technologies beyond narrow applications. Our proposed approach introduces a three-loop architecture, with this study focusing on the inverse loop. The framework aims to support R&D, facilitate real-time DT implementation, and remain interpretable to stakeholders beyond the immediate research community. Accordingly, the key contributions of this study can be summarized as:

1. As part of a broader, long-term research initiative, this study introduces the first published work on our proposed digital twin framework, which is structured around three interconnected loops: forward, inverse, and generative. Each loop serves a distinct function (see Section 2). Due to the technical complexity and validation requirements of each loop, this paper focuses specifically on the inverse loop, which represents the primary computational bottleneck for integrating sensor data into the digital twin.
2. We propose the Variational Digital Twin (VDT) concept, utilizing variational inference via a Bayesian last layer as a robust, efficient, and accurate method for real-time sensor data assimilation. This approach employs a batching mechanism, allowing the model to incrementally update itself as new data arrives, thereby improving its representation of the physical asset. The VDT framework also quantifies prediction uncertainty, enabling informed forward actions within the digital twin system.
3. The VDT framework is implemented across various neural network architectures that form the core of the digital twin. These include feedforward neural networks, recurrent neural networks, and physics-informed neural networks, demonstrating both compatibility with the approach and strong performance results.

**Nomenclature**

| | | | |
|---|---|---|---|
| AAL | Aided Active Learning | MCMC | Markov Chain Monte Carlo |
| AI | Artificial Intelligence | MHTGR | Modular High-Temperature Gas-cooled Reactor |
| CHF | Critical Heat Flux | ML | Machine Learning |
| CI | Confidence Interval | MSE | Mean Squared Error |
| DCC | Depressurized Conduction Cooldown | NN | Neural Network |
| DT | Digital Twin | PCC | Pressurized Conduction Cooldown |
| ELBO | Evidence Lower Bound | PINN | Physics-Informed Neural Network |
| FNN | Feedforward Neural Network | PSML | Power Systems Machine Learning |
| GNN | Graph Neural Network | RMSE | Root Mean Squared Error |
| GRU | Gated Recurrent Unit | RNN | Recurrent Neural Network |
| HTTF | High-Temperature Test Facility | RW | Random Walk |
| LOFC | Loss of Forced Convection | TF | Fluid Temperature |
| LR | Learning Rate | TS | Solid Temperature |
| LSTM | Long short-term memory | VDT | Variational Digital Twin |
| MAE | Mean Absolute Error | VI | Variational Inference |

4. The proposed concepts are validated through four distinct energy-related applications, spanning both steady-state and transient conditions. These include scenarios ranging from critical heat flux analysis to renewable energy systems, nuclear energy, and battery technologies. The results highlight the versatility and effectiveness of the VDT framework in addressing the previously identified gaps.

The structure of the paper is as follows: Section 2 outlines the conceptual framework of the DT developed in this study. Section 3 details the theoretical foundations and methodology guiding the data assimilation process within the variational DT approach. The energy-related applications and datasets used to validate the proposed concept are introduced in Section 4. Section 5 presents the outcomes from four applications, illustrating the effectiveness of the variational DT approach. Section 6 offers a discussion on the implications of applying variational methods in DTs, and Section 7 concludes the paper with a summary of key findings.

## 2. A Modular Concept of Digital Twins

This section introduces the foundational concept of our proposed Digital Twin (DT) framework as it is presented for the first time in the literature, while Section 3 elaborates on the mathematical formulation central to this study. The development of our DT model begins with the discrete-time dynamical system:

$$s_{t+1} = f(s_t, t), \tag{1}$$

where $s_t$ represents the current state of the DT (and the physical twin) at time step $t$, $f$ is the function governing the system's dynamics that the DT aims to learn (the primary goal of this paper), and $s_{t+1}$ is the predicted next state. Our objective is to ensure the DT can continuously update itself and remain synchronized with the physical asset, which streams data from onboard sensors.

We adopt a discrete-time formulation, rather than a continuous-time one, to reflect the practical limitations of data acquisition, processing, and model updating. Continuous, real-time updates are often infeasible in realistic settings due to processing overhead and communication latency. The proposed DT workflow, illustrated in Fig. 1, comprises three core operational loops: Forward, Inverse, and Generative. These loops are detailed next.

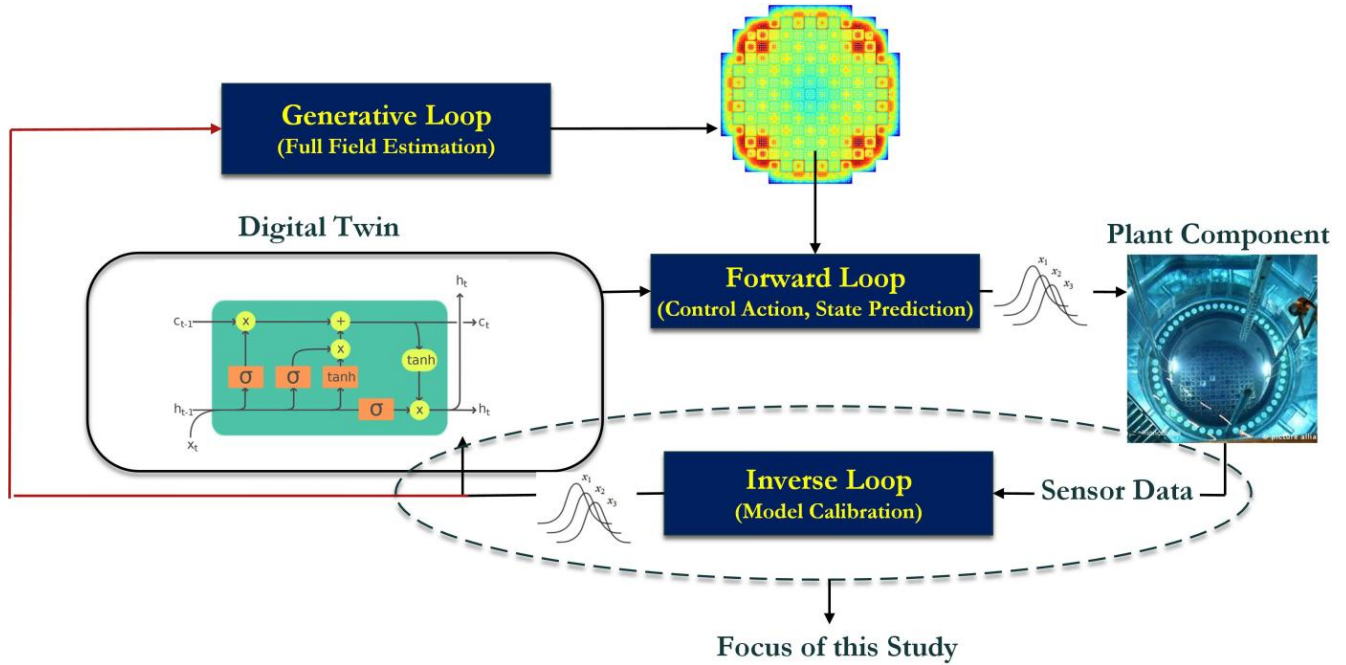

Figure 1: The conceptualization of the digital twin in this work.

## 2.1. Forward Loop

The forward loop is responsible for simulating or predicting the behavior of the physical asset based on the current state of the DT. Its functions include forecasting future system states and outputs, optimizing control inputs, and predicting potential failures or maintenance needs through trend analysis. As indicated in our literature review earlier, the forward loop is commonly referred to as the DT itself, encompassing a range of techniques from traditional modeling and simulation approaches [36] to data-driven surrogate models [37], such as recurrent neural networks [38], as well as methods rooted in control theory like model predictive control [32, 39], and advanced intelligent control strategies, including reinforcement learning [40, 41]. The primary objective of the forward loop is to enable direct, near-real-time decision-making and control based on the DT. Given that the concepts underpinning the forward loop—such as real-time control, process optimization, and state forecasting—are well-established, this study does not focus on it. However, the integration of the forward loop with the inverse loop, discussed below, will be addressed in future extensions of this work.

## 2.2. Inverse Loop

The inverse loop enables the DT to continuously refine its underlying models in real time by assimilating sensor data from the physical asset. This data assimilation process is essential for maintaining synchronization between the DT and the actual system as conditions evolve. To be effective, assimilation must occur with high frequency, low latency, and minimal computational overhead. However, this loop often presents a major computational bottleneck, particularly when the underlying models are complex and expensive to calibrate.

Traditionally, model calibration can be performed using direct optimization techniques to fit model parameters to minimize a cost function like model error [42, 26] or through Bayesian inverse methods supported by Markov

Chain Monte Carlo (MCMC) [43, 44]. While optimization methods risk overfitting due to the lack of a probabilistic framework, Bayesian MCMC approaches, though robust, are often prohibitively expensive for real-time applications.

In this work, we adopt a Bayesian framework to characterize uncertainty probabilistically. To overcome the computational limitations of MCMC, we introduce a novel variational inference-based data assimilation method. This technique is designed to strike a balance between inference accuracy and computational efficiency, enabling real-time calibration of the DT using practical CPU and GPU resources. The authors believe that the inverse loop represents a critical, underdeveloped component of DT systems and will therefore be the central focus of this study.

*2.3. Generative Loop*

The generative loop introduces an AI-driven component designed to reconstruct full-field system data in regions lacking direct sensor measurements. Acting as a bridge between the forward and inverse loops, it fuses high-fidelity simulation outputs with sparse sensor inputs to generate spatially complete and physically consistent representations of the system. A 3D generative model [45] may be implemented to infer missing field data, augment sensor and simulation outputs, and reconstruct signals in unsensed regions.

A novel generative framework will be developed in future work, going beyond traditional techniques such as interpolation, imputation, or sparse approximation methods like matching pursuits [46]. This approach aims to enhance the accuracy and coverage of the digital twin (DT) through efficient data augmentation and full-field estimation, thereby supporting more informed decision-making in the forward loop. While the DT can operate with just the forward and inverse loops, the generative loop serves as an optional enhancement. *This component is currently in early-stage development and will be detailed in future publications once a robust implementation is achieved.*

## 3. Methodology

*3.1. Variational Inference and Neural Architectures*

Variational inference (VI) provides a powerful Bayesian alternative to conventional point-estimate learning by approximating complex, intractable posterior distributions with simpler, tractable ones [47]. In the context of deep learning, this framework enables networks to learn distributions over parameters, rather than fixed weights, thereby encoding model uncertainty directly into the learning process. This is particularly valuable for applications involving noisy, high-dimensional, or evolving data—such as in scientific and engineering domains where uncertainty estimation is critical.

Concretely, let $\boldsymbol{\theta}$ denote the collection of neural network weights and biases, and let $D = \{(\mathbf{x}^{(n)}, y^{(n)})\}^{N}_{n=1}$ be the dataset, where N is the number of samples. We begin by placing a prior $p(\boldsymbol{\theta})$ over the network parameters. The posterior $p(\boldsymbol{\theta} \mid D)$ is generally intractable. In VI, we introduce a variational distribution $q(\boldsymbol{\theta}; \boldsymbol{\phi})$ (parameterized by variational parameters $\boldsymbol{\phi}$) to approximate the true posterior. We then maximize the evidence lower bound (ELBO):

$$\text{ELBO}(\boldsymbol{\phi}) = \mathbb{E}_{q(\boldsymbol{\theta};\boldsymbol{\phi})}[\log p(D \mid \boldsymbol{\theta})] - \beta \times \text{KL}(q(\boldsymbol{\theta}; \boldsymbol{\phi}) \parallel p(\boldsymbol{\theta})). \quad (2)$$

Maximizing the ELBO jointly encourages (i) high likelihood of the observed data under sampled parameters and (ii) closeness of the variational $q$ to the prior. In practice, we choose $q(\boldsymbol{\theta}; \boldsymbol{\phi})$ to factorize over weights and biases (mean-field assumption) and employ the reparameterization trick [48] to obtain low-variance gradient estimates of the ELBO with respect to $\boldsymbol{\phi}$. By replacing deterministic layers with variational layers, we endow the neural network with a posterior distribution over its parameters. During training, each forward pass samples $\boldsymbol{\theta} \sim q(\boldsymbol{\theta}; \boldsymbol{\phi})$, computes the likelihood term $\log p(D \mid \boldsymbol{\theta})$, and accumulates the KL regularization. $\beta$ is a constant used to ensure relative closeness of the magnitudes of the reconstruction loss and KL divergence such that one does not entirely dominate the loss function. We find $\beta$ of $10^{-4}$ to generalize well across the selected applications. At inference time, multiple samples from $q(\boldsymbol{\theta}; \boldsymbol{\phi})$ yield predictive distributions $\hat{y} \sim p(y \mid \mathbf{x}, \boldsymbol{\theta})$, from which one can extract both a mean prediction and uncertainty bounds (e.g., credible intervals).

When applied to neural networks, variational methods can be used to introduce a stochastic layer where weights are sampled from learned posterior distributions during training and inference. These sampled predictions yield both a mean prediction and an uncertainty estimate, forming the foundation for probabilistic reasoning, risk-aware decision making, and active learning strategies. Importantly, the Bayesian nature of these models allows for principled updating as new data becomes available, aligning naturally with the continuous learning requirements of digital twin systems. We quantify uncertainty via construction of 95% confidence intervals (CI). For each test input we draw $S$ stochastic forward passes $y^{(1)}, y^{(2)}, \ldots, y^{(S)}$ to compute the predictive mean,

$$\bar{y} = \frac{1}{S} \sum_{s=1}^{S} y^{(s)} \tag{3}$$

and the 95% confidence lower and upper bounds,

$$y_{\text{lower}} = \text{Quantile}(\{y^{(s)}\}, 2.5\%), \qquad y_{\text{upper}} = \text{Quantile}(\{y^{(s)}\}, 97.5\%) \tag{4}$$

In this work, we integrate variational last layers [49] into deep neural architectures—including feedforward networks, recurrent networks, and physics-informed models—to create variational digital twins (VDTs). We leverage variational inference primarily to (i) quantify predictive uncertainty in digital twin outputs, (ii) allow for Bayesian updating as new data arrive, and (iii) evaluate the digital twin's confidence in its forecasts. These models have the potential not only deliver competitive predictive performance but also generate confidence intervals that evolve as the model encounters new information, enabling both adaptive forecasting and uncertainty quantification across diverse physical systems.

Feedforward neural networks (FNNs) apply a succession of affine maps followed by nonlinear activations to transform an input vector $\mathbf{x} \in \mathbb{R}^d$ into an output $\mathbf{y}$. Writing $L$ for the number of layers and $\sigma(\cdot)$ for the activation, the governing relations are

$$\mathbf{a}^{(0)} = \mathbf{x}, \qquad \mathbf{a}^{(\ell)} = \sigma(\mathbf{W}^{(\ell)} \mathbf{a}^{(\ell-1)} + \mathbf{b}^{(\ell)}), \qquad \ell = 1, \ldots, L, \tag{5}$$

$$\mathbf{y} = \mathbf{a}^{(L)}, \tag{6}$$

where $\mathbf{W}^{(\ell)}$ and $\mathbf{b}^{(\ell)}$ are the trainable weights and biases of layer $\ell$.

Recurrent neural networks (RNNs) maintain a hidden state $\mathbf{h}_t$ that evolves with each time-step input $\mathbf{x}_t$. The vanilla RNN is described by

$$\mathbf{h}_t = \phi(\mathbf{W}_h \mathbf{h}_{t-1} + \mathbf{W}_x \mathbf{x}_t + \mathbf{b}), \qquad \mathbf{y}_t = \mathbf{W}_y \mathbf{h}_t + \mathbf{c}, \tag{7}$$

where $\phi(\cdot)$ is a nonlinear activation. Two gated variants, introduced to resolve the vanishing/exploding gradient problem [50] of vanilla RNNs, are used here.

Long short-term memory (LSTM) [51] cells introduce memory $c_t$ and three gates:

$$\mathbf{i}_t = \sigma(\mathbf{W}_i \mathbf{x}_t + \mathbf{U}_i \mathbf{h}_{t-1} + \mathbf{b}_i), \tag{8}$$

$$\mathbf{f}_t = \sigma(\mathbf{W}_f \mathbf{x}_t + \mathbf{U}_f \mathbf{h}_{t-1} + \mathbf{b}_f), \tag{9}$$

$$\mathbf{o}_t = \sigma(\mathbf{W}_o \mathbf{x}_t + \mathbf{U}_o \mathbf{h}_{t-1} + \mathbf{b}_o), \tag{10}$$

$$\tilde{\mathbf{c}}_t = \tanh(\mathbf{W}_c \mathbf{x}_t + \mathbf{U}_c \mathbf{h}_{t-1} + \mathbf{b}_c), \tag{11}$$

$$\mathbf{c}_t = \mathbf{f}_t \odot \mathbf{c}_{t-1} + \mathbf{i}_t \odot \tilde{\mathbf{c}}_t, \tag{12}$$

$$\mathbf{h}_t = \mathbf{o}_t \odot \tanh(\mathbf{c}_t). \tag{13}$$

The input gate $\mathbf{i}_t$ uses the current input and previous hidden state to modulate how much of the candidate update $\tilde{\mathbf{c}}_t$ enters the cell memory. The forget gate $\mathbf{f}_t$ similarly determines what fraction of the prior cell state $\mathbf{c}_{t-1}$ to retain. Finally, the output gate $\mathbf{o}_t$ controls how much of the updated cell state $\mathbf{c}_t$ (after a tanh nonlinearity) is revealed as the new hidden state $\mathbf{h}_t$.

Gated recurrent units (GRUs) [52, 53] achieve a similar effect with two gates:

$$\mathbf{z}_t = \sigma(\mathbf{W}_z \mathbf{x}_t + \mathbf{U}_z \mathbf{h}_{t-1} + \mathbf{b}_z), \tag{14}$$

$$\mathbf{r}_t = \sigma(\mathbf{W}_r \mathbf{x}_t + \mathbf{U}_r \mathbf{h}_{t-1} + \mathbf{b}_r), \tag{15}$$

$$\tilde{\mathbf{h}}_t = \tanh(\mathbf{W}_h \mathbf{x}_t + \mathbf{U}_h (\mathbf{r}_t \odot \mathbf{h}_{t-1}) + \mathbf{b}_h), \tag{16}$$

$$\mathbf{h}_t = (1 - \mathbf{z}_t) \odot \mathbf{h}_{t-1} + \mathbf{z}_t \odot \tilde{\mathbf{h}}_t. \tag{17}$$

The update gate $\mathbf{z}_t$ determines how much of the previous hidden state $\mathbf{h}_{t-1}$ versus the new candidate $\tilde{\mathbf{h}}_t$ is carried forward, while the reset gate $\mathbf{r}_t$ controls how much of $\mathbf{h}_{t-1}$ is used in computing that candidate. The candidate hidden state $\tilde{\mathbf{h}}_t$ is produced by applying a tanh nonlinearity to a combination of the current input and the gated previous state ($\mathbf{r}_t \odot \mathbf{h}_{t-1}$). Finally, the new hidden state $\mathbf{h}_t$ is a convex interpolation between $\mathbf{h}_{t-1}$ and $\tilde{\mathbf{h}}_t$ weighted by $\mathbf{z}_t$.

Physics-informed neural networks (PINNs) [54, 55] embed a learned ODE residual into the loss. State evolution is approximated by an explicit-Euler update $x_{n+1}$ and an output map $\hat{y}_n$.

$$x_{n+1} = x_n + f_\theta(x_n, u_n) \Delta t \tag{18}$$

$$\hat{y}_n = h_\theta(x_n) \tag{19}$$

Given the output $\hat{y}_n$, the composite training objective can be defined as the sum of the data and physics-based loss,

$$L_{\text{data}} = \frac{1}{N} \sum_{n=1}^{N} \|\hat{y}_n - y_n\|^2 \tag{20}$$

$$L_{\text{phys}} = \frac{1}{N-1} \sum_{n=1}^{N-1} \left\| \frac{x_{n+1} - x_n}{\Delta t} - f_\theta(x_n, u_n) \right\|^2 \tag{21}$$

$$L = L_{\text{data}} + \lambda L_{\text{phys}} \tag{22}$$

where $x_n \in \mathbb{R}^d$ denotes the system's state at time step $n$, and $u_n$ is the known control or input; the function $f_\theta(x_n, u_n)$ is a parameterized vector field modeling the state's time-derivative, with $\Delta t$ the step size in the explicit-Euler update for $x_{n+1}$. The map $h_\theta(x_n)$ is a separate network decoding the latent state $x_n$ into the predicted observation $\hat{y}_n$, which is compared against the true data $y_n$ in $L_{\text{data}}$. The physics loss $L_{\text{phys}}$ enforces that the finite-difference derivative $(x_{n+1} - x_n)/\Delta t$ matches $f_\theta(x_n, u_n)$, and the scalar $\lambda \geq 0$ balances data fidelity against ODE consistency in the total loss $L$.

Throughout this study, variational last layers are appended to these three backbones (FNN, LSTM/GRU, PINN) to endow the digital twin with calibrated Bayesian uncertainty while preserving the specialised strengths of each architecture.

### 3.2. Active Learning for Sensor Sampling

Active learning (AL) helps to optimize resource utilization by prioritizing experiments/simulations that are expected to yield the most impactful results [56]. In this work, we employed an aided AL (AAL) approach as a way to sample sensor/experimental data that improves digital twin predictions while minimizing data requirements. Unlike conventional AL, which selects the samples based on uncertainty or error directly from the entire pool, AAL first applies random sampling to create a smaller candidate pool before applying the query strategy. Training starts with 10 samples, adding 20 samples per iteration, selected from a pre-randomized batch of 500. Fifty independent trials/experiments were conducted to ensure statistical robustness, with sampling continuing until the model reached a target predictive performance threshold (e.g., $R^2$ = 0.98). This two-step sampling prevents premature bias, risk of overfitting and promotes better generalization. Our preliminary analysis indicated that AAL was less greedy in sample selection compared to applying the query strategy directly to the available data pool, leading to a better DT model. AAL in this study is demonstrated on the case study of critical heat flux (CHF) prediction described in Section 4.1.

### 3.3. Digital Twin Update Loop

To evaluate the applicability of the proposed VDT approach to time-series prediction problems in the energy sector, we implemented RNN models for forecasting solar and wind power generation using meteorological inputs from the PSML (Power Systems Machine Learning) dataset described in Section 4.2. The digital twin was designed to assimilate temporally resolved data and generate multi-output predictions over a rolling forecast horizon.

A discrete-time VDT is implemented as follows: at each time step, the model received a sequence of meteorological inputs spanning the prior 12 minutes, encoded them via a deep LSTM/GRU architecture, and predicted the power generation levels for solar and wind energy as outputs. The models were pre-trained on one month of initial data and subsequently updated using new training data at the start of each subsequent two-month period (termed a "session"), simulating periodic DT recalibration.

To emulate real-world constraints such as limited memory and computational bandwidth, the DT was incrementally retrained using only the most recent month's data, discarding older data to mimic edge-deployed models. After each session, the model's performance was evaluated on a subsequent one-month test period using $R^2$, mean absolute error (MAE), and root mean squared error (RMSE) as metrics. Model parameters were saved after each update and used to initialize the next session, thereby enabling temporal continuity and knowledge retention across updates. The approach adopted for this session-based training is given in Algorithm 1.

---

**Algorithm 1** Session-Based Variational Digital Twin Updating for Power Grid Forecasting Applications (PSML)

---

**Require:** Time-series dataset D with meteorological inputs and power outputs
**Require:** Sequence length $L$, session window size $W$
**Require:** Initialized GRU model $M_0$ with parameters $\theta_0$
**Require:** Loss function L, learning rate $\eta$, number of epochs $E$
1: Preprocess D: resample, impute, and scale features
2: Construct sequence pairs $(\mathbf{X}, \mathbf{y})$ using window size $L$
3: Split $(\mathbf{X}, \mathbf{y})$ into train/test sessions of size $W$
4: **for** each session $t = 0, 1, \ldots, T$ **do**
5:     Define training set $T_t$ and test set $V_t$
6:     **if** $t > 0$ **then**
7:         Load previous model weights: $\theta_t \leftarrow \theta_{t-1}$
8:     Initialize optimizer with $\theta_t$
9:     **for** epoch $e = 1$ to $E$ **do**
10:         **for** each batch $(x_b, y_b) \in T_t$ **do**
11:             Forward pass: $\hat{y}_b \leftarrow M_t(x_b)$
12:             Compute loss: $\ell \leftarrow L(y_b, \hat{y}_b)$
13:             Backpropagate and update: $\theta_t \leftarrow \theta_t - \eta \nabla \ell$
14:     Evaluate $M_t$ on test set $V_t$ and store metrics
15:     Save model weights: $\theta_t$
16: **return** Metric evolution across sessions

---

This session-based updating strategy allows the VDT to remain synchronized with the physical system while adapting to seasonal shifts, weather trends, or infrastructural changes in the energy grid. Importantly, this approach mirrors how DT might operate in deployment: continuously updated, temporally local, and performance-tracked in real time.

*3.4. Sensor Data Assimilation*

The High Temperature Test Facility (HTTF) experiment dataset described in detail in Section 4.3 consists of high-resolution thermal data collected during a depressurized conduction cooldown (DCC) scenario over 1.9 days, sampled at 2 Hz. Each of the internal reactor vessel sensors measures either solid temperature (TS) or fluid temperature (TF), yielding 329,067 time steps per sensor. For computational efficiency and noise reduction, we downsample the data to 30-second intervals before processing. Our objective is to generate two long 1D sequences—one for TS and one for TF—that preserve temporal structure and enhance the learning capacity of sequential models like GRUs.

A key insight is that the ordering of sensor time series in the final concatenated signal significantly influences the presence of learnable temporal patterns (e.g., seasonality, drift, periodicity). If sensors are concatenated arbitrarily, the resulting signal may appear chaotic or trendless, inhibiting sequence modeling and hence affecting DT learning. To address this, we develop a structured sensor concatenation algorithm guided by the average temperature readings of each sensor. The core idea is to group sensors into bins based on their average signal levels and then interleave their order in a way that promotes smooth transitions and quasi-periodic patterns across the concatenated sequence. Each sensor's time series is averaged, and all sensors are grouped into 10 quantile-based bins according to their mean value. Within each bin, sensors are sorted in ascending order by their mean temperature, creating local sequences that reflect increasing thermal gradients. The final global ordering of sensors is constructed iteratively. On even-numbered passes, we select the lowest-mean sensor remaining from each bin; on odd-numbered passes, we select the highest. This zig-zag pattern prevents monotonic buildup and fosters artificial periodicity in the final signal. The ordered sensor indices are then used to concatenate their respective time series end-to-end, forming a single 1D signal for TS and TF. This process is detailed in Algorithm 2.

---

**Algorithm 2** Sensor Concatenation Strategy for Time-Series Modeling

---

**Require:** Set of sensor time series $\{X_i\}_{i=1}^{M}$ with $T_i$ time steps each
**Ensure:** Concatenated signal $Y$ with learnable temporal patterns
1: **for** each sensor $i = 1$ to $M$ **do**
2:     Compute mean temperature: $\mu_i \leftarrow \frac{1}{T_i} \sum_{t=1}^{T_i} x_i(t)$
3: Bin sensor indices into $K = 10$ quantile groups using $\{\mu_i\}$
4: **for** each bin $B_k$ **do**
5:     Sort sensors in ascending order of $\mu_i$ to get ordered list $L_k$
6: Initialize global order list $O \leftarrow []$
7: Initialize cycle counter $c \leftarrow 0$
8: **while** any list $L_k$ is nonempty **do**
9:     **for** each bin index $k = 0$ to $9$ **do**
10:         **if** $L_k$ is nonempty **then**
11:             **if** $c \bmod 2 == 0$ **then**
12:                 Append front element of $L_k$ to $O$ and remove it
13:             **else**
14:                 Append back element of $L_k$ to $O$ and remove it
15:     Increment cycle counter: $c \leftarrow c + 1$
16: Concatenate time series in $O$ to form $Y$
17: **return** $Y$

---

This signal processing strategy helps introduce structure into the signal that recurrent models can exploit, particularly when no inherent sensor spatial topology is available. The resulting data stream more closely resembles structured temporal phenomena, which are known to improve RNN learning dynamics.

Following sensor preprocessing and structured concatenation, we implement a sequential learning framework to predict reactor vessel thermal behavior. The objective was to forecast both TS and TF trajectories using a single unified model. The concatenated sensor signals were split into training, validation, and test sets, with the model trained to infer unseen temperature profiles in the test set from patterns learned in the training set.

*3.5. Variational Physics-Informed Digital Twin*

Lastly, we apply our VDT framework to model the degradation dynamics of Li-ion batteries using data from NASA's RW3 cell under randomized discharging profiles, described in Section 4.4. The model builds upon BattNN [57], a physics-informed neural network developed to estimate battery voltage from variable current profiles by integrating three data-driven subnetworks with a physics-based equivalent circuit model (ECM) of the battery, as displayed in Figure 2.

The three networks are used to estimate the internal battery parameters. The first network maps the charge, $q_b$, to the state-of-charge (SOC). The other networks map the SOC to the battery voltage ($V_b$) and the resistance induced by surface overpotentials ($R_{sp}$). These components are embedded within a physics-based core that enforces charge conservation and voltage laws via coupled difference equations over internal states $q_b$, $q_{sp}$, and $q_s$. The final predicted terminal voltage $V$ results from a combination of estimated subcomponent voltages and currents, grounded in electrical circuit analogies.

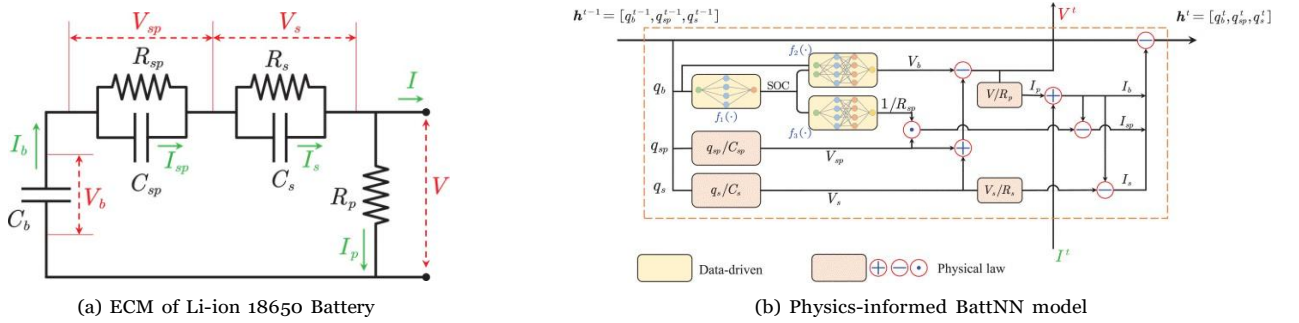

(a) ECM of Li-ion 18650 Battery

(b) Physics-informed BattNN model

Figure 2: Battery ECM and BattNN model developed by [57].

To adapt this framework for variational inference, we augment each of the networks with variational last layers, enabling uncertainty quantification and Bayesian updating of network parameters. In contrast to the original BattNN training approach, which shuffles discharging profiles across time, we organize training and testing data chronologically. This mimics a realistic deployment scenario for DT, where models must continuously learn from past cycles to predict future behavior. The model is fine-tuned iteratively using a rolling window scheme. At each iteration, a block of discharges is used for training, followed by uncertainty-aware predictions on the next block. New measurements are incorporated as they become available, enabling robust continual learning over a six-month horizon.

This approach allows the DT to forecast future discharge curves with credible intervals, making it ideal for use cases where understanding the uncertainty in voltage predictions is essential—such as early warning systems for battery failure, state-of-health estimation, and adaptive battery management. Further, we explore whether our model can adapt over time without necessarily modifying the physical parameters in the governing equations.

### 3.6. Model Error and Performance Metrics

Let $\{y_i\}_{i=1}^N$ denote the ground-truth observations and $\{\hat{y}_i\}_{i=1}^N$ the corresponding model predictions.
Mean Absolute Error (MAE) measures the average magnitude of the residuals without regard to sign.

$$\text{MAE} = \frac{1}{N} \sum_{i=1}^{N} |y_i - \hat{y}_i|.$$

Mean Squared Error (MSE) penalizes larger deviations more severely by squaring the residuals.

$$\text{MSE} = \frac{1}{N} \sum_{i=1}^{N} (y_i - \hat{y}_i)^2.$$

Root Mean Squared Error (RMSE) is the square root of MSE, restoring the units of the original variable:

$$\text{RMSE} = \sqrt{\text{MSE}}.$$

The Coefficient of Determination ($R^2$) quantifies the fraction of variance in the data explained by the model:

$$R^2 = 1 - \frac{\sum_{i=1}^{N}(y_i - \hat{y}_i)^2}{\sum_{i=1}^{N}(y_i - \bar{y})^2}, \qquad \bar{y} = \frac{1}{N}\sum_{i=1}^{N} y_i.$$

Lower values of MAE, MSE, and RMSE indicate better predictive accuracy. RMSE is more sensitive to outliers because of the squaring operation, while MAE provides a robust, scale-preserving error measure. $R^2$ ranges from $(-\infty, 1]$; a value of 1 corresponds to perfect predictions, whereas $R^2 \leq 0$ signifies performance no better than predicting the mean $\bar{y}$. Each metric is used to give a balanced view of both average and variance-weighted errors across the test sets.

## 4. Data and Applications

### 4.1. Critical Heat Flux (CHF)

The first application of this study utilizes the CHF dataset published by the OECD Nuclear Energy Agency in 2024 [58], under the supervision of the U.S. Nuclear Regulatory Commission (NRC), for predicting CHF, which is a critical safety parameter for nuclear reactors. CHF is the point at which a phase change in the coolant (typically from liquid to vapor) causes the formation of a vapor blanket or film on the heated surface (e.g., fuel cladding). This phenomenon severely reduces the coolant's ability to remove heat, resulting in a rapid increase in surface temperature — a condition known as departure from nucleate boiling.

The dataset contains over 21 thousand CHF measurements on water-cooled vertical tubes, compiled from 59 different sources. Five parameters: pressure ($P$), mass flux ($G$), and inlet temperature ($T_{\text{in}}$) and 2 geometric parameters: channel diameter ($D$) and heated length ($L$) were chosen as input features for the DT model, while CHF is considered the target output. The span of the parameters in the CHF dataset is summarized in Table 1.

Table 1: Parameter spans of the CHF dataset used in this study [58]

| Variable | Channel Diameter (mm) | Heated Length (m) | Pressure (kPa) | Mass Flux (kg/m²s) | Inlet Temperature (°C) | CHF (kW/m²) |
|---|---|---|---|---|---|---|
| Minimum | 2.39 | 0.07 | 100.0 | 17.7 | 9.0 | 130.0 |
| Maximum | 16.0 | 15.0 | 20,000.0 | 7,712.0 | 353.62 | 13,345.0 |
| Mode | 8.0 | 3.0 | 9,800.0 | 1,499.0 | 279.86 | 1500.0 |

### 4.2. Power Systems Machine Learning (PSML)

This study incorporates the PSML dataset developed by Zheng et al. [59] for forecasting renewable power generation from meteorological features. The dataset captures real-world and synthetic time-series data across various spatiotemporal scales, including minute-level and millisecond-level resolutions, with data collected over a three-year period (2018–2021) from 66 regions in the United States. Zone 1 of the California Independent System Operator (CAISO) is selected for this study. Eight meteorological parameters serve as DT model inputs—Diffuse Horizontal Irradiance (DHI), Direct Normal Irradiance (DNI), Global Horizontal Irradiance (GHI), Dew Point, Solar Zenith Angle, Wind Speed, Relative Humidity, and Temperature—with the VDT model predicting two outputs: wind and solar power generation. All features and targets, excluding the solar zenith angle, as it follows the same exact pattern each day, are pictured over the 3-year time period in Figure 3.

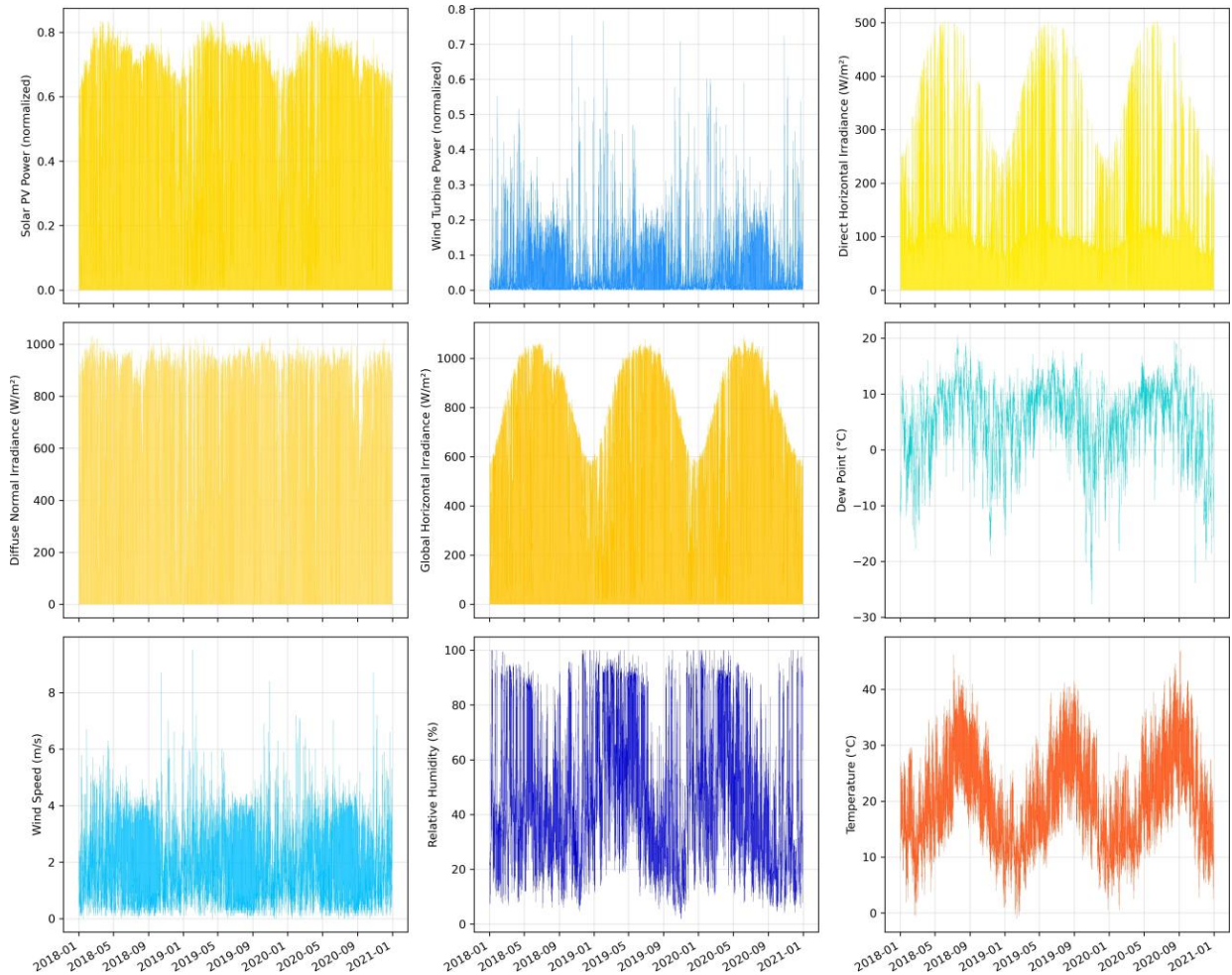

Figure 3: PSML CAISO Zone 1 dataset (excluding solar zenith angle) over a 3 year period from 2018 to 2021. Targets for prediction are solar and wind generation, displayed in first two plots in the top left.

*4.3. High Temperature Test Facility (HTTF) at Oregon State University (OSU)*

The HTTF at OSU is an integral, large-scale experimental platform developed to replicate key thermal–hydraulic behaviors of Modular High-Temperature Gas-cooled Reactors (MHTGRs) [60, 61, 62]. The facility utilizes helium as a working fluid and employs electrical heaters capable of delivering approximately 2.2 MWt. The system is highly instrumented and capable of reproducing transient scenarios such as Loss of Forced Convection (LOFC), Pressurized Conduction Cooldown (PCC), and Depressurized Conduction Cooldown (DCC) events. In this study, time-resolved sensor data from the HTTF DCC experiment is used for state estimation and uncertainty quantifi-cation during transient thermal-hydraulic events. The DCC experiment takes place over 1.9 days, and 155 solid and fluid temperature sensors (310 total) demonstrate the thermal evolution taking place in various parts of the reactor vessel, pictured in Figure 4.

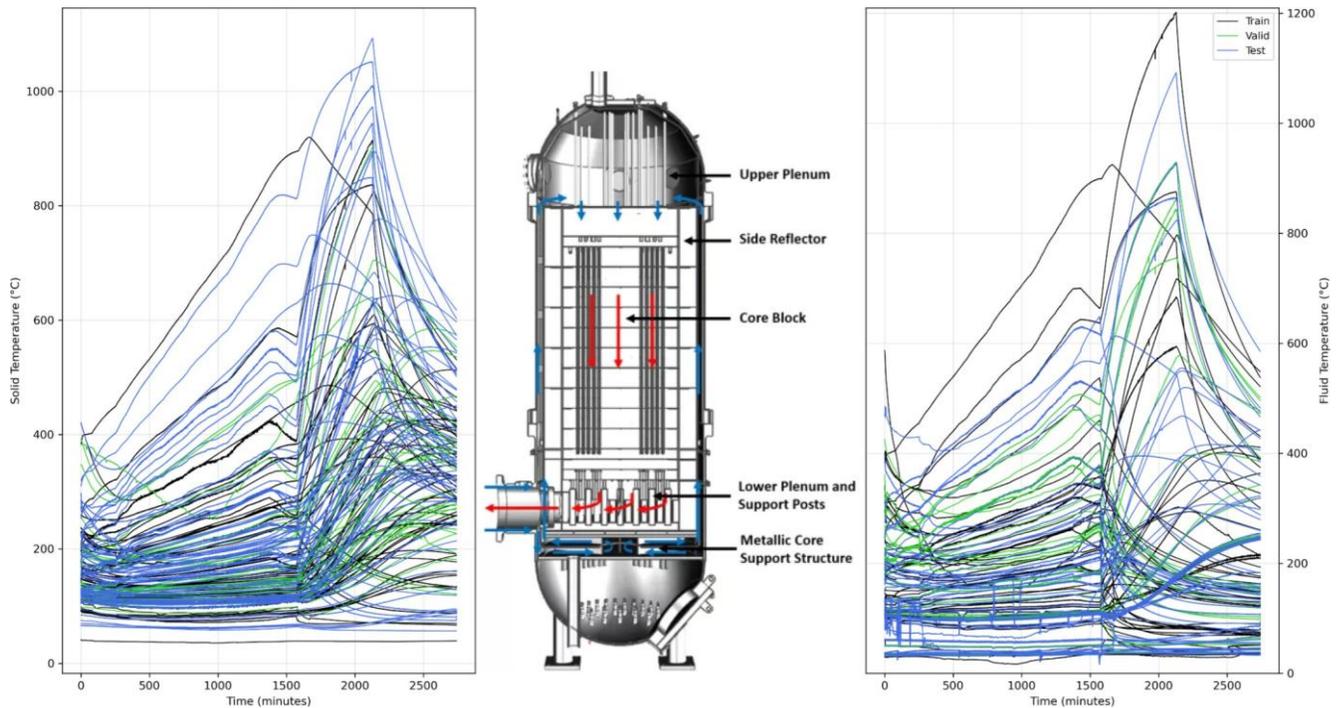

Figure 4: OSU HTTF schematic and visualization of recorded solid (left) and fluid (right) temperature profiles over nearly 2 days.

*4.4. NASA Li-ion Battery Degradation*

This work uses the public NASA Random Walk (RW) Li-ion battery dataset, which features long-term operational data from four 18650 Li-ion cells [63]. The batteries were cycled repeatedly using a random walk (RW) current discharge protocol, wherein each cell was charged to 4.2V and discharged to 3.2V under a randomized sequence of currents ranging from 0.5 A to 4 A. We focus on battery RW3 for demonstration purposes and aim to model the future voltage trajectory of the cell under arbitrary current discharge profiles. The current profiles serve as input features, while the voltage time series constitutes the output target for the VDT model. Randomly selected discharges are displayed in chronological order in Figure 5.

The final summary of the four applications used in this work to test the proposed VDT framework is given in Table 2.

Table 2: Summary of the datasets used in this work

| **Dataset Name** | **Field of Interest** | **Type** | **Reference** |
|---|---|---|---|
| Critical Heat Flux (CHF) | Thermal Hydraulics | Static | [58] |
| Power Systems Machine Learning (PSML) | Renewable Energy Forecasting | Time-series | [59] |
| High Temperature Test Facility (HTTF) | Nuclear Reactor Safety | Time-series | [62] |
| NASA Li-ion Battery Degradation | Battery Health Monitoring | Time-series | [57] |

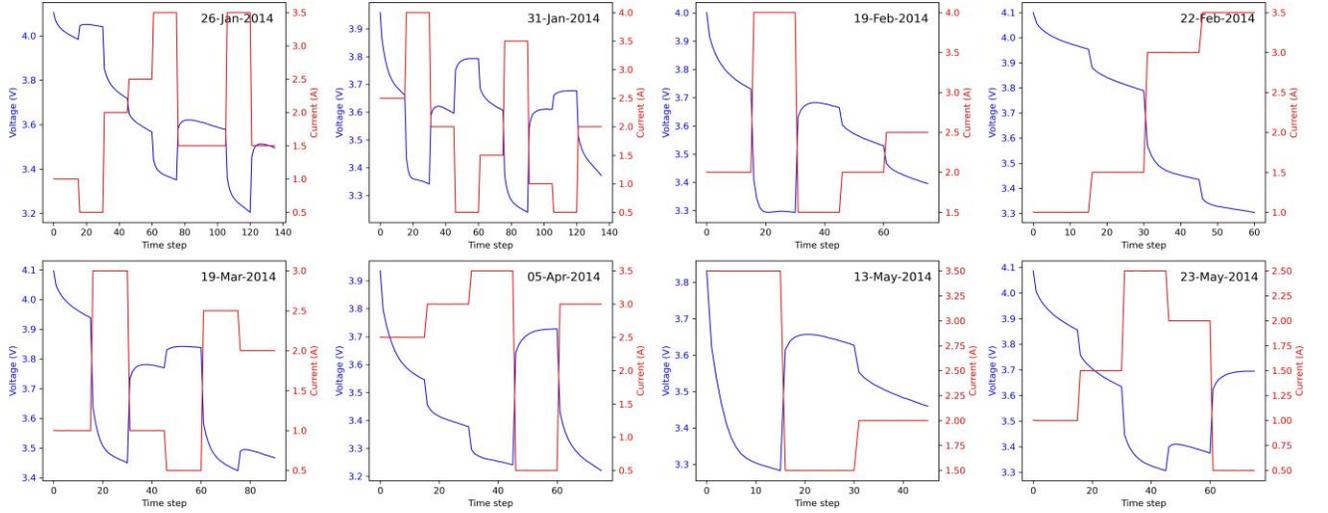

Figure 5: Eight randomly selected random walk current-voltage profiles from the NASA RW3 Li-ion battery degradation dataset.

## 5. Results

Table 3 summarises the core architectures and training settings used in each application prior to the performance analyses that follow. Across all four case studies we adopt a consistent strategy: a deterministic backbone—feed-forward network for CHF, single- or multi-layer RNNs for PSML and HTTF, and three physics-motivated MLP subnetworks for the Li-ion battery model— all augmented by a single variational Bayesian linear layer. Learning rate (LR), hidden-layer widths, and training duration are listed along with any other relevant hyperparameters. Each architecture was modestly tuned using a simple gird search to keep model complexity low and comparable across applications. The uniform yet lightweight design of each model enables us to highlight how variational inference and the proposed VDT updating, rather than extensive architectural engineering, can drive improvements in data efficiency, model robustivity, and uncertainty calibration.

Table 3: Architectural summary and training settings for the variational models. Layers with subscript $\sigma$ are variational layers.

| Application | Model Type | Architecture | Hyperparameters |
|---|---|---|---|
| CHF | VFNN | $5 \rightarrow 100 \rightarrow 100 \rightarrow 100_\sigma \rightarrow 1$ | LR = $10^{-3}$<br>100 epochs<br>Batch size = variable (8-64)<br>Weight decay = $10^{-5}$ |
| PSML | VLSTM/VGRU | $2 \rightarrow 35 \rightarrow 35_{RNN} \rightarrow 35 \rightarrow 35_\sigma \rightarrow 2$ | LR = $10^{-3}$<br>50 epochs<br>Batch size = 512 |
| HTTF | VLSTM/VGRU | $2 \rightarrow (48 \rightarrow 64 \rightarrow 32)_{RNN} \rightarrow 32_\sigma \rightarrow 2$ | LR = $1.8 \times 10^{-4}$<br>50 epochs<br>Batch size = 256 |
| NASA Li-ion | VBattNN | $q_b \rightarrow SOC$: $1 \rightarrow 4 \rightarrow 4_\sigma \rightarrow 1$<br>$q_b, SOC \rightarrow V_b$: $2\text{-}8 \rightarrow 8 \rightarrow 8_\sigma \rightarrow 1$<br>$SOC \rightarrow 1/R_{sp}$: $1 \rightarrow 8 \rightarrow 8 \rightarrow 4_\sigma \rightarrow 1$ | LR = $2 \times 10^{-2}$<br>1000 epochs<br>Batch size = 10<br>Weight decay = $5 \times 10^{-4}$<br>Half LR @ epochs 100, 500 |

*5.1. Aided Active Learning for Critical Heat Flux Digital Twin*

In this application, the VDT model is represented by a VFNN model that receives the data in batches sampled using aided active learning (AAL). The VDT inputs are pressure ($P$), mass flux ($G$), inlet temperature ($T_{in}$), channel diameter ($D$), and heated length ($L$), while CHF is acting as the DT output. Figure 6 compares the VFNN model performance for CHF prediction using AAL and random sampling strategies described in Section 3.2. The AAL approach, based on uncertainty sampling, shows a steeper rise in $R^2$ during early iterations by prioritizing the most informative samples to assimilate into the VDT, thereby accelerating model improvement. Throughout the sampling process, AAL consistently outperforms random sampling, achieving higher $R^2$ scores with reduced variability. Complementarily, Figure 7 illustrates the cumulative training time required by both strategies, revealing that AAL achieves the high predictive performance significantly faster than random sampling. This highlights not only the improved data efficiency of AAL but also its substantial advantage in computational time savings.

As shown in Figure 7, we intentionally highlight that the training cost of the DT can increase exponentially in both AAL and random sampling scenarios, due to the growing data pool over time, which forces the VDT to retrain on increasingly larger datasets. This issue will be addressed in the next application, where we implement session-based DT updates and reload the model weights instead of retraining from scratch.

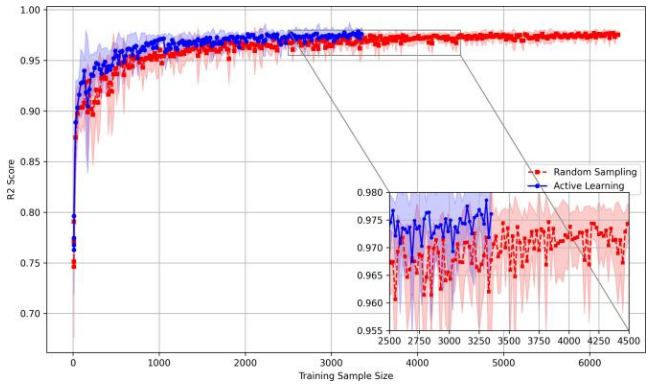

Figure 6: Mean $R^2$ score of CHF with increasing training samples for the VDT model (VFNN), using uncertainty sampling.

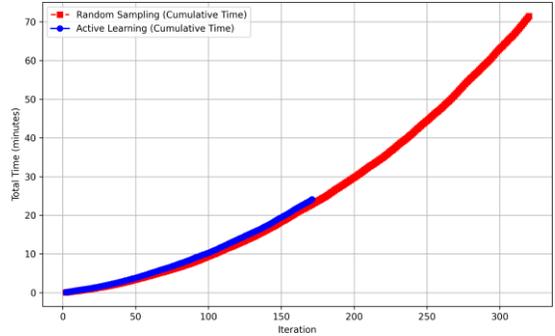

Figure 7: Cumulative VDT training time for AAL and random sampling to reach target predictive performance of CHF.

Table 4 summarizes the training sample size and corresponding training time required to achieve an $R^2$ score of 0.98 (acceptable threshold for this application) for random sampling and active learning strategies. AAL achieves the target accuracy with nearly half the number of samples compared to random sampling (3420 vs. 6440 samples), demonstrating a 47% reduction in data assimilation requirements for the DT. Furthermore, the time needed for training is significantly lower for AAL, requiring only 24 minutes compared to 72 minutes for random sampling. The tests performed in this case were conducted on a single NVIDIA RTX 6000 Ada GPU.

Table 4: Training sample size and time required to achieve an $R^2$ score of 0.98 by the VDT for the CHF dataset

| Method | Training Size | Time (min) |
| --- | --- | --- |
| Random Sampling | 6440 | 72 |
| Aided Active Learning | 3420 | 24 |

*5.2. PSML Digital Twin Updating*

This section presents the training and performance outcomes of VRNN models benchmarked against their traditional, deterministic counterparts as well as the application of Algorithm 1 for VDT updating. Table 5 and Figure 8 demonstrate that the VRNNs maintain predictive accuracy comparable to deterministic models in forecasting both solar and wind power generation. Despite introducing stochasticity through variational inference, the VRNNs yield nearly identical performance across $R^2$, MAE, and RMSE. Crucially, the variational models also produce uncertainty bounds around each forecast as seen in the insets of Figure 8, allowing access to confidence in predictions and identification of periods of elevated uncertainty. These uncertainty estimates are especially valuable in real-time grid operations, where probabilistic forecasts can inform reserve allocation, load balancing, and decision support under uncertainty.

Table 5: Performance metrics for power-generation forecasting on the PSML dataset

| Model | Solar | | | Wind | | |
|---|---|---|---|---|---|---|
| | MAE | RMSE | $R^2$ | MAE | RMSE | $R^2$ |
| GRU | 0.029404 | 0.061822 | 0.946125 | 0.003242 | 0.007017 | 0.991292 |
| VGRU | 0.028031 | 0.056625 | 0.954802 | 0.003032 | 0.005054 | 0.995484 |
| LSTM | 0.028243 | 0.058560 | 0.951659 | 0.002908 | 0.005414 | 0.994816 |
| VLSTM | 0.027638 | 0.057239 | 0.953817 | 0.003034 | 0.004972 | 0.995629 |

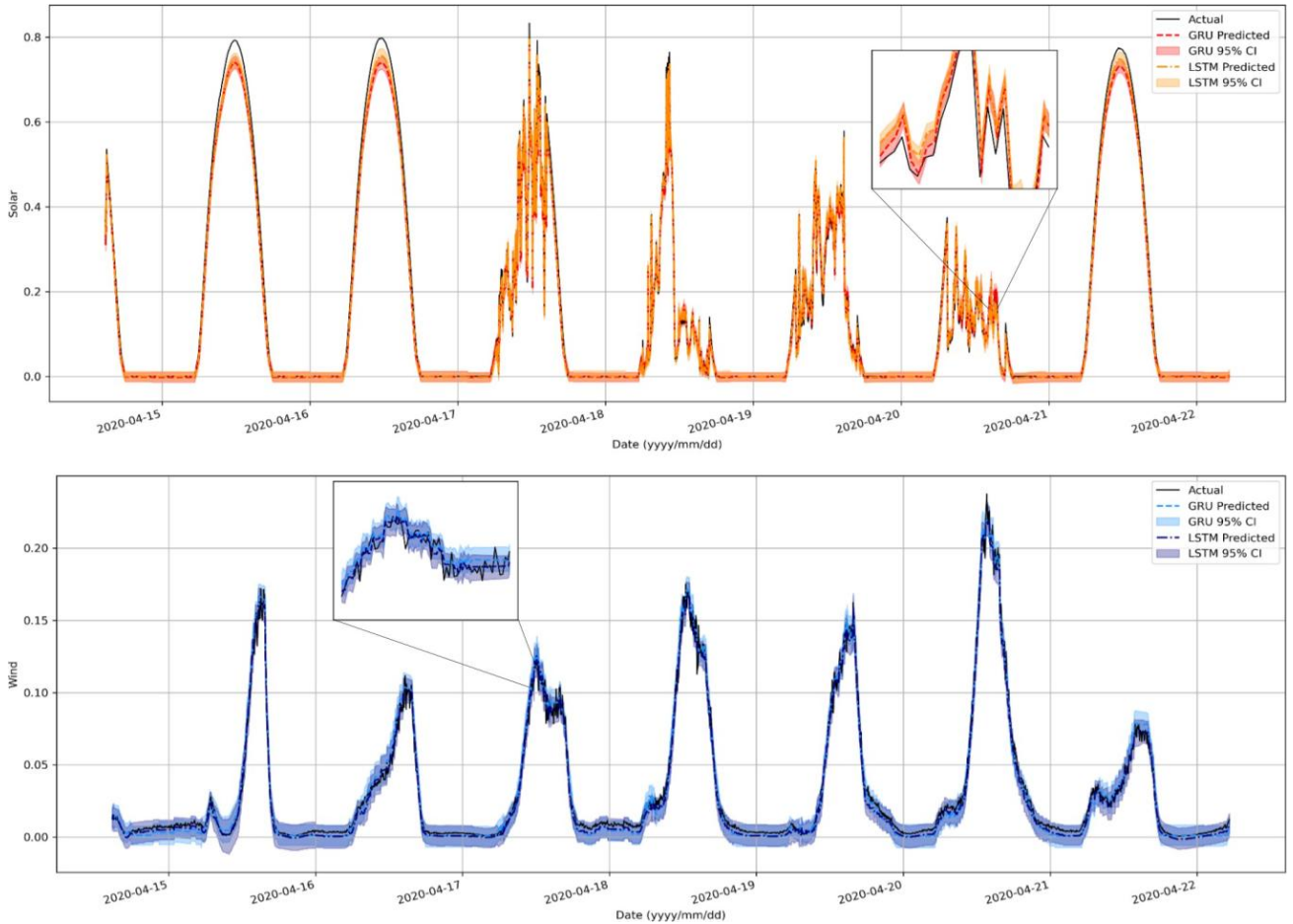

Figure 8: Comparison of VLSTM/VGRU predictions over 1 week after single-shot training.

However, this only shows the ability of VRNNs to compete with deterministic models, not their ability to adapt and improve in a DT setting. For this, we implement Algorithm 1 to emulate a situation where the model initially only has access to one month's worth of power grid data, attempts to predict the next month, and then, once the forecasted month has passed, the model consumes that month of new data to update its internal state and learn any new or emerging dynamics due to changes in weather, grid infrastructure, or other environmental factors. Once retrained, the model then attempts to predict the next month's worth of solar and wind generation, and the process is repeated. This iterative loop was used to cover the nearly 3 years, or 1.57 million data points, contained in the PSML dataset. A resulting 33 retraining sessions and subsequent predictions show that the VRNN model not only achieves impressive predictive accuracy using just the first month of data, but also manages to maintain high accuracy as it is updated over time to reflect changes to CAISO Zone 1's solar and wind power generation systems. The model remains resilient to distributional drift in input features and retains reasonably low error rates across seasonal transitions, all while continuously producing locally-calibrated uncertainty estimates that track forecast confidence throughout deployment.

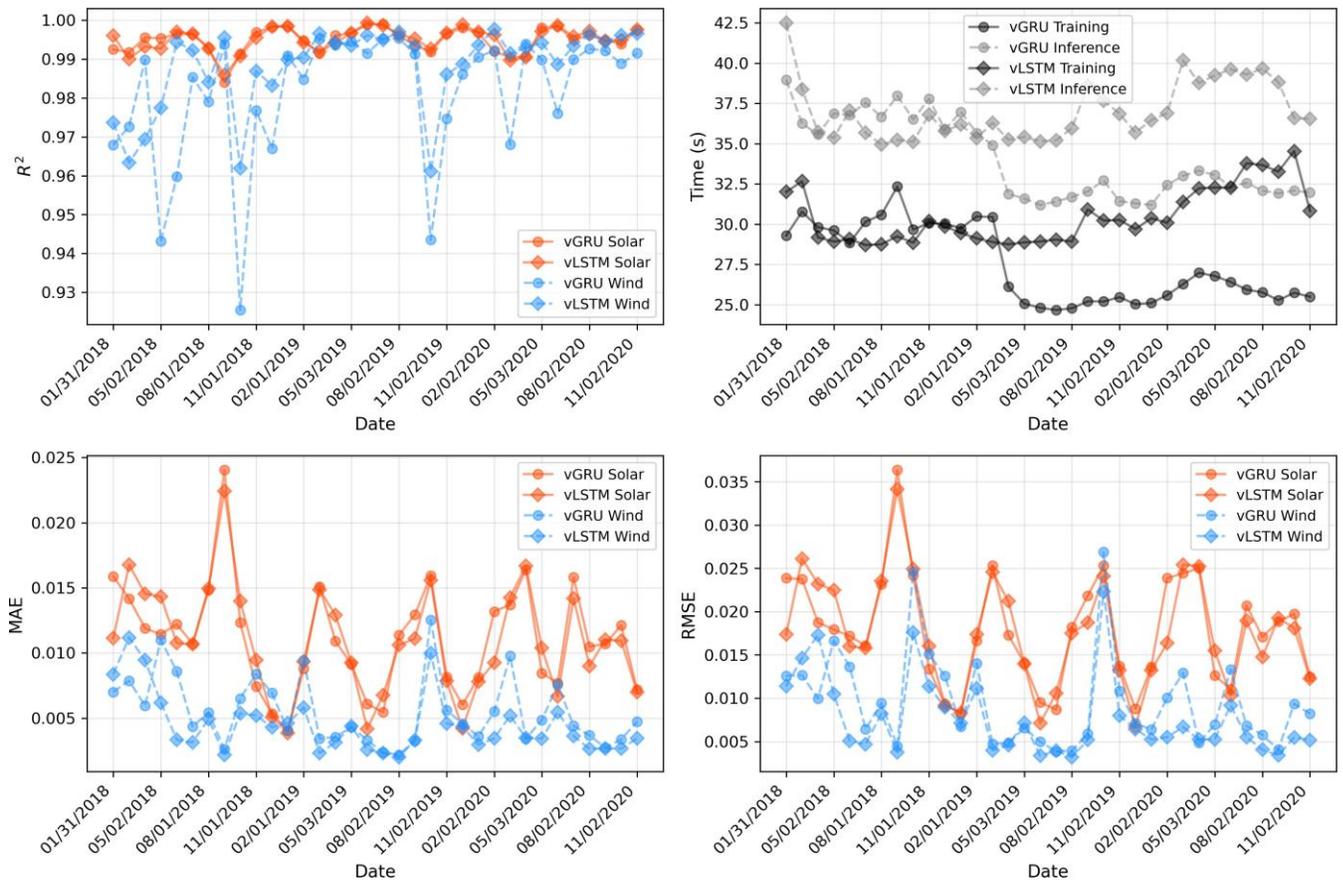

Figure 9: Session-wise comparison of VGRU (circles) vs. VLSTM (diamonds) for solar (solid, red) and wind (dashed, blue); top-right shows training/inference times.

A key advantage of the proposed VDT framework lies in its computational scalability and update efficiency. Rather than requiring access to the entire 1.57 million-point PSML dataset for training, the VRNN model operates in a rolling update loop, retraining only on the most recent month of data before making forward predictions. Each retraining session is lightweight and requires less than 55 seconds on NVIDIA GeForce RTX 4060 Ti, enabling practical, near-real-time deployment on relatively modest hardware when compared with the demands of many modern machine learning algorithms. This avoids the exponential growth in training time and memory footprint that we observed before in Figure 7, which would occur if the model were continuously retrained on an ever-growing dataset.

While VGRU demonstrates the expected reduction in training time over VLSTM, VLSTM outperforms in terms of accuracy, especially for wind, throughout the 3-year period. Regardless, $R^2$ trends in Figure 9 show that the quality and consistency of wind power prediction by both variational models is worse than solar prediction. Specifically, the $R^2$ for wind in the static training case was significantly higher, and the opposite is true for solar. Wind power's high variability and rare regime shifts are best learned from a static model trained on the full historical record, since it retains long-memory of extremes and distributional patterns. However, solar power, dominated by slowly evolving seasonal cycles, benefits more from iterative retraining on recent data, which keeps the model aligned with current weather patterns. *Thus, static training preserves wind's long-term variability, while rolling windows provide the "fresh" seasonal context that solar forecasting requires*. Regardless, despite never seeing more than a fraction of the data at any one time, the model sustains high predictive accuracy across nearly three years of operation. Furthermore, the top-right section of Figure 9 shows that the training and inference times for VDT models range between 25 and 45 seconds — an impressive performance considering that each session uses a commodity GPU and contains minute-level data spanning an entire month, reinforcing the model's real-time applicability. This demonstrates the digital twin's ability to generalize from short-term context, adapt to evolving system behavior, and scale gracefully with asset age. Such properties will be essential for long-term deployment of data-driven DT for energy systems.

*5.3. HTTF Sensor Assimilation and Imputation*

The HTTF dataset presents an opportunity to evaluate the ability of another RNN-based VDT to model the collective dynamics of a large array of sensors as opposed to the isolated measurements contained in the PSML dataset. The unique problem of assimilating data from multiple sensors is tackled using Algorithm 2 in which sensors measuring the same physical quantity are concatenated into a single, 1D time-series. The iterative approach to combining sensor signals based on their relative mean optimizes the resulting signal for periodic structure and stability. As seen in Figure 10, the concatenated signal will enable recurrent models to exploit the synthetic patterns despite the lack of any obvious or inherent topology initially present in the dataset. It is important to note that the number of sensors available in this facility (155 for both TS and TF) is relatively small compared to its physical size, making it impractical to leverage their spatial locations within a graph neural network (GNN) structure. GNNs also require the explicit encoding of geometric and spatial information, making it application-specific and thus lacking the qualities of a generalizable digital twin framework. For this reason, we adopted a sensor-agnostic concatenation approach instead.

LSTM and GRU neural networks were employed, designed to capture temporal dependencies across the composite signal. The model received sequences from both TS and TF channels and was trained to jointly predict the next temperature values. Training was conducted over 50 epochs using the Adam optimizer with mean squared error (MSE) loss, and model performance was evaluated on a held-out test set. To quantify the model's robustness and practical utility, we performed a sensitivity analysis by progressively reducing the number of sensors available during training. This analysis aimed to identify the minimum viable sensor subset required to maintain predictive accuracy on unseen test signals. The outcome of this experiment could also be used to inform future decisions on sensor placement and redundancy in reactor monitoring systems.

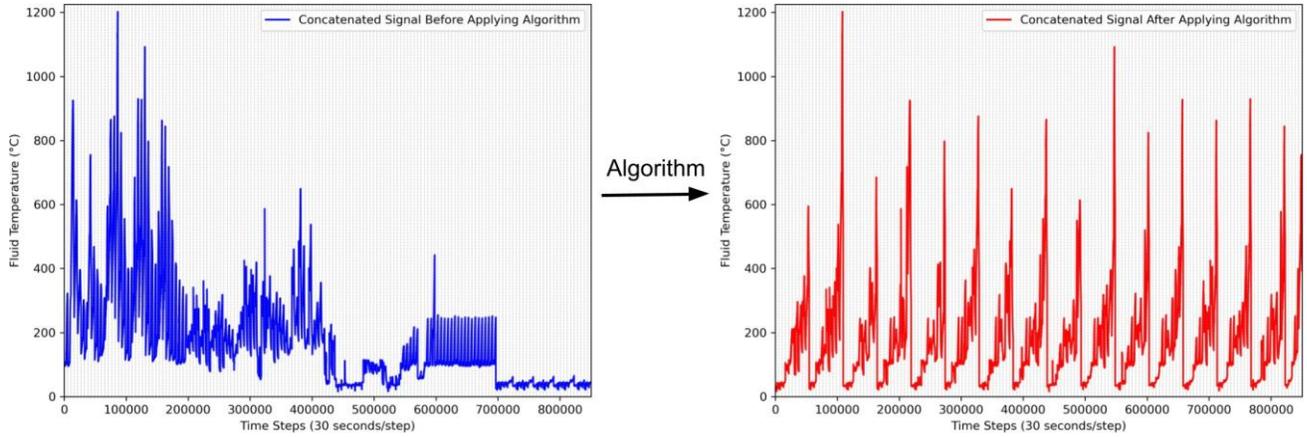

Figure 10: Full 1D concatenation of fluid temperature sensor signals before and after applying Algorithm 2 for stacking sensor data to enhance sequential VDT performance. Vertical gray lines indicate the boundary where one sensor signal ends and the next begins.

Once trained, the VLSTM model demonstrated a strong ability to track the complex thermal evolution of the reactor vessel throughout the depressurized conduction cooldown (DCC) event. From Figure 11, predictions of both solid temperature (TS) and fluid temperature (TF) profiles closely followed the true sensor measurements over the full 1.9-day test sequence. Forecasts captured both long-term temperature trends and local transient fluctuations in thermal behavior across the reactor vessel domain. The probabilistic nature of the variational model readily allows for generation of uncertainty bounds alongside temperature predictions. These intervals adaptively widen during periods of increased thermal instability and rapid change and thus offer a principled estimate of model confidence at each point in time. However, Table 6 shows that the variational models, while able to provide uncertainty quantification with their predictions, display slightly degraded accuracy. Interestingly, VGRU and VLSTM appear to find distinct local minimas for TS and TF during training as their MAE is nearly perfectly inverted. Despite the small degradation in performance, the ability to quantify uncertainty is particularly valuable in the scope of safety-critical nuclear applications where understanding the reliability of digital twin forecasts *can and should* influence data-driven operator decisions and protocols.

Training and inference are also longer, but this is more than expected from a hybrid Bayesian neural architecture and stays well within practical time limits given that the models are predicting nearly 2 days worth of operation in only a few hundred seconds. Beyond training, we compare model inference times via inference rate, or how many predictions the model can make per second. Given that RNNs necessarily condition their predictions on the information from previous time steps, there is little opportunity to utilize the parallelization capabilities of GPUs for inference. However, variational models sample the same prediction many times to evaluate uncertainty, and thus inference rate can be significantly increased via parallelization. Inference still naturally requires more time overall compared to deterministic models, but the ability to parallelize sampling across the GPU keeps inference time within a few hundred seconds despite repeated sampling for each prediction.

In the accompanying sensitivity analysis, we progressively reduced the number of available sensors during training to simulate scenarios of partial instrumentation or sensor failure and evaluate the convergence of model accuracy and uncertainty quantification. Results in Figure 12 show that the models remained robust even when trained on as few as 30 (50%) of the original sensors, but predictions and uncertainty begin to diverge once $\sim 40$ sensors have been removed. This suggests that the VDT is capable of reconstructing thermal field dynamics from a reduced set of observations, but there is also a necessary amount of information required from the physical asset to obtain reliable performance and uncertainty. Moreover, Figure 12 suggests VLSTM is significantly more sensitive to the reduction in training data with increasingly sharp changes in accuracy as more sensors are removed. VGRU not only maintains higher predictive accuracy than VLSTM but also is much more robust against the changes available sensor data. This resilience likely stems from the GRU's inherently simpler gating structure and reduced parameter count, which allows the VGRU to learn and propagate the most salient temporal features even when input dimensionality shrinks. When combined with variational layers, this architecture enforces stronger regularization and produces smoother, more reliable uncertainty estimates under data scarcity. In contrast, the more complex VLSTM, with its separate input, output, and forget gates, requires more data to tune its larger parameter space and therefore shows sharper performance degradation as sensors are removed. This is a promising capability for VGRU in real-world systems where sensor coverage may be limited or degrade over time, especially in systems where access to sensored areas is difficult and/or infrequent. The HTTF study illustrates how a VDT can learn to model physically meaningful behavior in a highly complex, sensor-dense system. A deployable DT must be capable of adapting to temporal dynamics, but also to limitations in sensor availability, while delivering both accurate forecasts and interpretable uncertainty measures.

Table 6: HTTF temperature prediction performance metrics for GRU, LSTM, VGRU, and VLSTM

| Model | Output | $R^2$ | MAE (°C) | MAPE | RMSE (°C) | Train Time (s) | Infer Rate (pred./s) |
|---|---|---|---|---|---|---|---|
| GRU | TS | 0.9981 | 3.1902 | 1.5483 | 7.9551 | 252.0917 | $2.38 \times 10^5$ |
| | TF | 0.9958 | 5.9329 | 4.9220 | 9.7476 | | |
| LSTM | TS | 0.9963 | 3.6903 | 1.6514 | 11.3122 | 260.2081 | $2.28 \times 10^5$ |
| | TF | 0.9961 | 6.0366 | 6.5509 | 9.4317 | | |
| VGRU | TS | 0.9955 | 10.4948 | 6.3826 | 12.4349 | 396.1374 | $5.90 \times 10^5$ |
| | TF | 0.9961 | 6.6770 | 4.2884 | 9.4093 | | |
| VLSTM | TS | 0.9962 | 5.2490 | 3.2448 | 11.4196 | 420.2136 | $5.63 \times 10^5$ |
| | TF | 0.9886 | 10.9640 | 8.4715 | 16.1311 | | |

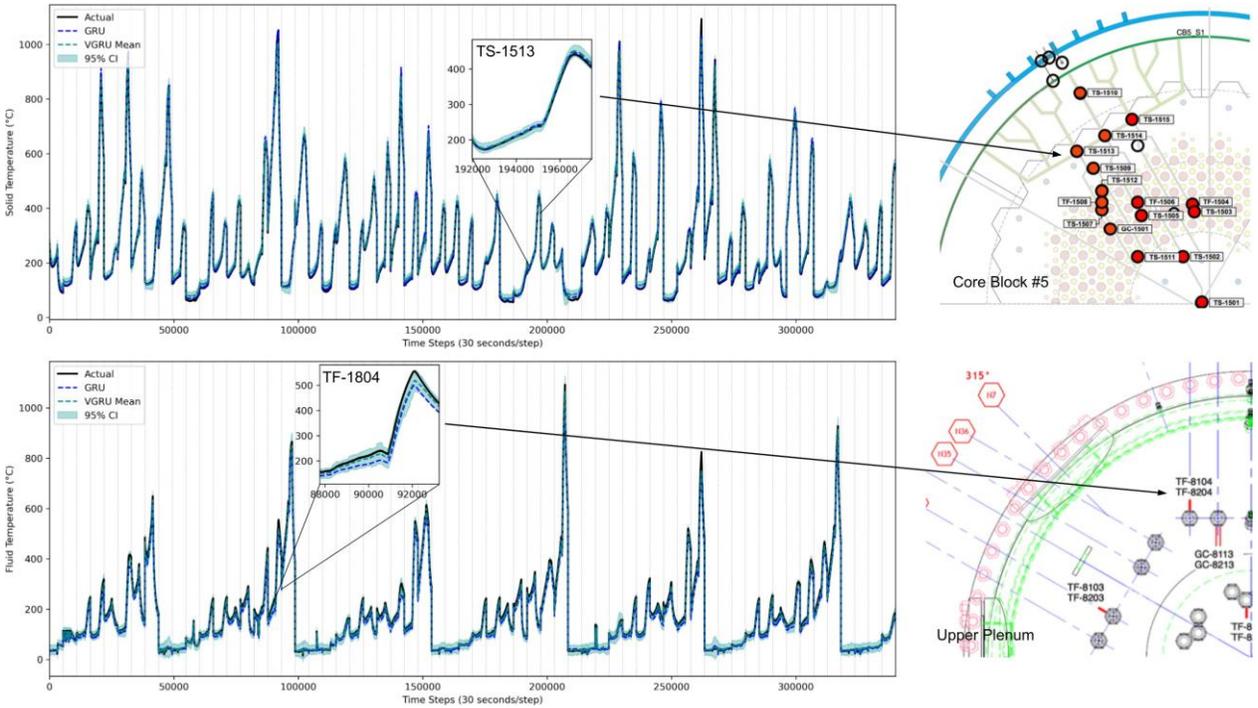

Figure 11: VGRU predictions for 62 solid and fluid temperature thermocouples in the unseen test set. Vertical gray lines indicate the boundary where one sensor signal ends and the next begins. Inset plots show a single sensor prediction and its corresponding location in the reactor on the right.

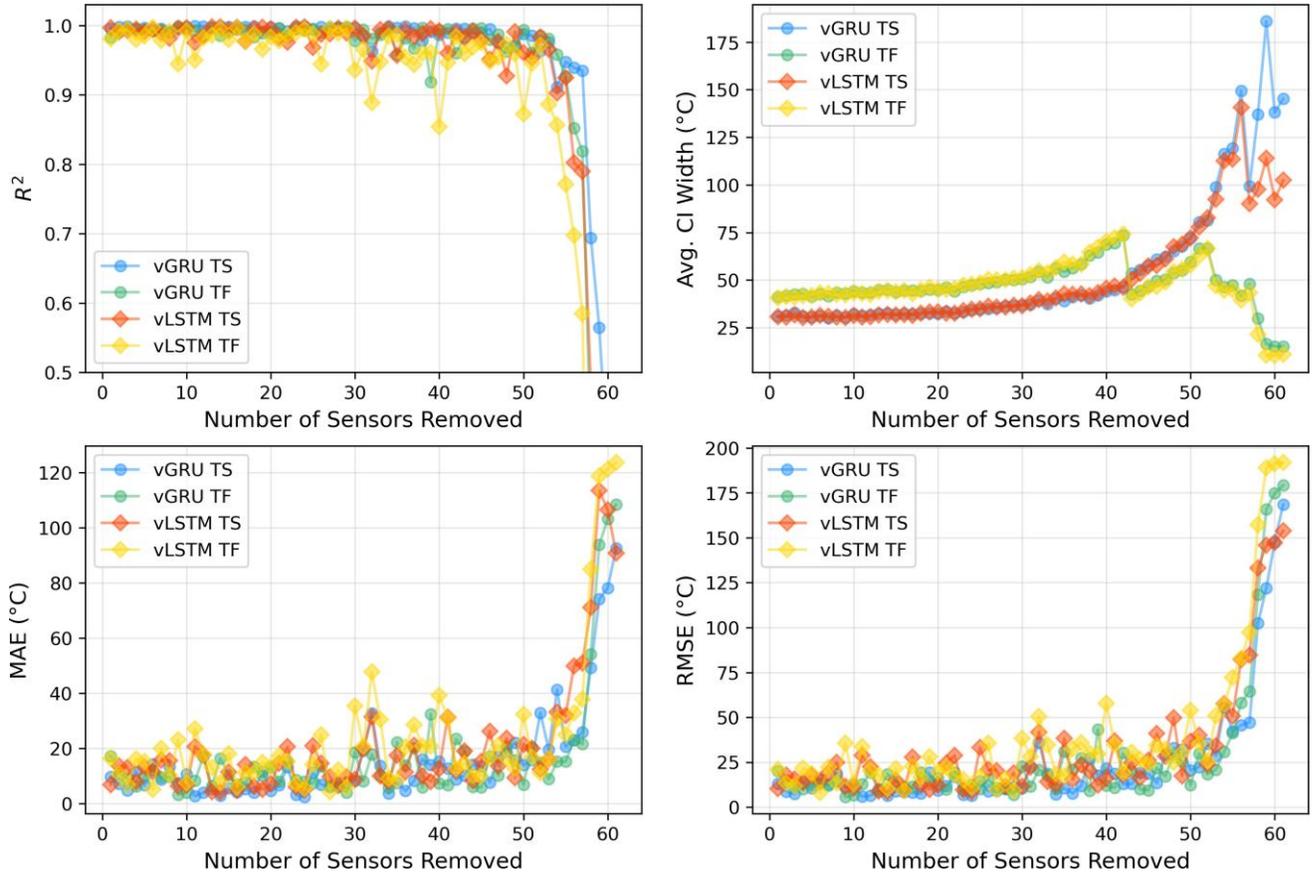

Figure 12: Degradation of VRNN model predictions as sensor signals are removed from the train set, with 60 removed indicating training on only one signal. Performance metrics and average confidence interval width converge with increasing access to sensor data.

As we increase the number of sensors used for training, the average predictive uncertainty of both VGRU and VLSTM converges steadily and begins to plateau as given in Figure 12. With 30 sensors (50%) are included, the uncertainty estimates for both TS and TF outputs converge to nearly identical values. This indicates that once a moderate level of spatial coverage is reached, the variational inference machinery in both architectures governs uncertainty calibration, outweighing differences in their gating mechanisms. Thus, although VGRU is more robust under extreme sensor reduction, both VGRU and VLSTM deliver comparable confidence once sufficient data are available.

*5.4. Li-ion Battery Degradation Digital Twin*

In this section, we present the performance evaluation of our Li-ion battery VDT under both static and iterative training regimes. We first summarize key metrics in Table 7 and then illustrate temporal prediction behavior and uncertainty evolution across the cell's lifecycle in Figures 13 and 14. Table 7 demonstrates that the VDT update

procedure—where the model is fine-tuned every ten discharges using ten new runs—yields markedly superior performance compared to any one-off, static training regimen. Specifically, the iterative VDT achieves an MSE of 0.00454 V², a MAPE of 1.41 %, and an RMSE of 0.0637 V. In contrast, static models trained on 30, 60, 150, 300, or 500 discharges exhibit MSE values in the range 0.022–0.027 V², MAPE values of 3.7–4.0 %, and RMSE values of 0.14–0.15 V. These results confirm that, despite accessing the same or greater total data at once, the static approaches cannot match the adaptability of a model that continuously incorporates fresh observations.

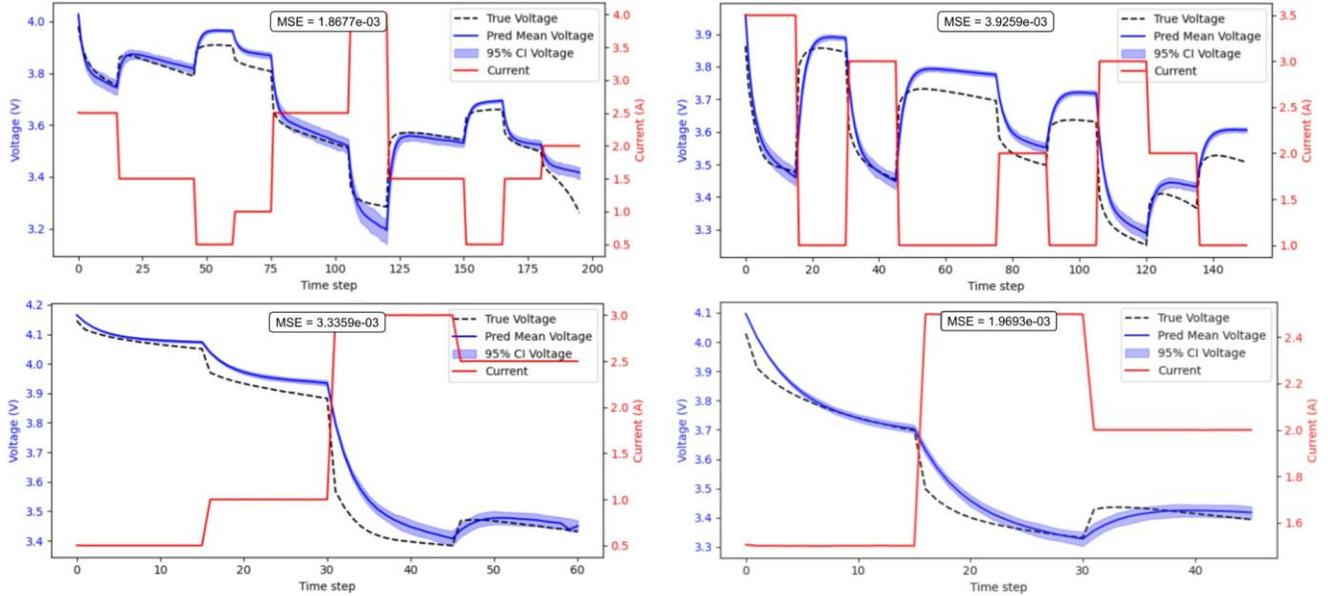

Figure 13: Discharge voltage predictions (blue) via iteratively updated vBattNN and corresponding current (red) profiles. Each taken from distant points in battery lifetime to demonstrate model adaptability to degradation of the physical battery, with lifetime increasing in the plots left to right and top to bottom.

Figure 13 presents representative discharge segments drawn from early life, mid-life, and near end-of-life to illustrate the VDT's evolving accuracy and confidence intervals over battery degradation. Early in the cycle, the VDT produces near-perfect voltage reconstructions (MSE ∼ $1.87 \times 10^{-3}$ V²) with narrowly bounded 95 % credible intervals, even in the presence of changing current profiles. As degradation progresses, mean predictions remain closely aligned with the true voltage traces, and the uncertainty bands widen appropriately to encompass the regions of high model discrepancy. Near end-of-life, the model continues to track the accelerated voltage drop (MSE ∼ $1.97 \times 10^{-3}$ V²) while providing well-calibrated uncertainty bounds, thereby demonstrating both accuracy and reliable uncertainty quantification throughout the cell's lifespan.

Figure 14 extends this analysis over all 750 + chronological discharges, plotting per-discharge MSE and MAPE for each static training size (colored traces) alongside the VDT rolling update procedure (gray markers). All static models exhibit pronounced error accumulation as the battery ages, with smaller training sets displaying the steepest growth and greatest scatter. By contrast, the VDT error also grows but at a substantially slower rate, maintaining a low, nearly flat baseline throughout most of the degradation trajectory. This slower error growth confirms that periodic variational updating effectively tracks the cell's evolving behavior far better than static training alone.

With BattNN being a physics-informed neural network, these findings suggest that, although iterative variational retraining significantly enhances temporal adaptability, further reductions in error growth might be achieved by periodically updating the fixed physical parameters within the underlying circuit model. Incorporating occasional recalibration of those physical parameters alongside variational layer updates could allow the DT to fully compensate for both data-driven and physics-driven drifts, thereby sustaining performance over the entire operational horizon.

Table 7: vBattNN Metrics by Training Procedure for the Li-Ion Battery Dataset

| Training Procedure | MSE | MAPE | RMSE |
| --- | --- | --- | --- |
| VDT Update, 10 runs/update | 0.004535 | 0.014055 | 0.063663 |
| Static, 150 runs | 0.025122 | 0.039195 | 0.147181 |
| Static, 30 runs | 0.027116 | 0.039892 | 0.149652 |
| Static, 300 runs | 0.022707 | 0.037708 | 0.142274 |
| Static, 500 runs | 0.024531 | 0.040522 | 0.152288 |
| Static, 60 runs | 0.027245 | 0.040367 | 0.151290 |

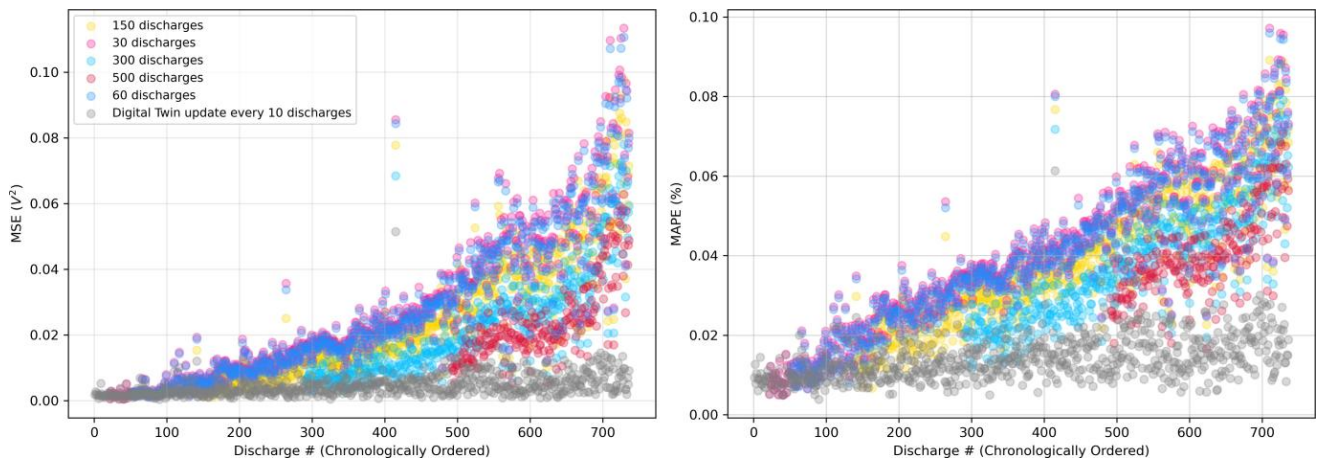

Figure 14: Comparison of vBattNN performance metrics, MSE (left) and MAPE (right) for varying train set sizes and the proposed VDT rolling update procedure.

One apparent weakness of using variational inference for this dataset seems to emerge somewhat randomly when training on small amounts of data. While in the other three applications the rolling predictions are stable, large deviations from the mean can be infrequently observed following a training update. The predicted mean voltage profile remains useful with low overall MSE, but the infrequent deviations suggest that variational-based models may struggle to calibrate uncertainty when not trained on sufficient data. However, given the coarseness of this dataset (200 seconds per time step and approximately 150 steps per discharge profile) this weakness should be isolated to data-scarce environments.

## 6. Discussions and Implications

The four case studies collectively demonstrate how a VDT architecture can close several long-standing gaps identified in the energy systems literature: sensor data assimilation, real-time model deployability, and the absence of uncertainty quantification. By embedding lightweight variational layers at the output of otherwise conventional neural networks, the proposed framework delivers Bayesian posterior estimates at negligible computational overhead, enabling the twin to reason about its own confidence while it learns. This section draws out the practical consequences of that capability for researchers and industrial practitioners, with an emphasis on (i) accelerated data acquisition via active learning, (ii) robust assimilation of heterogeneous sensor fields, and (iii) scalable, life-cycle forecasting under evolving operating conditions.

*6.1. Active learning with Uncertainty Quantification*

In critical-heat-flux (CHF) modelling, empirical tests remain expensive and time-consuming; thus, reducing the number of experiments without compromising predictive fidelity of the DT is a priority for both regulators and fuel-vendor R&D groups. Our variational feed-forward network (VFNN) turns its predictive variance into an *information score* that selects the next most informative experiments. Relative to random sampling, the aided active-learning (AAL) strategy achieved a target $R^2$=0.98 using 47 % fewer data points and in one-third the wall-clock time. For laboratory programmes constrained by test-loop availability or irradiation budgets, this translates directly into cost savings and shorter design cycles. More broadly, the result confirms that uncertainty estimates produced by simple variational last layers are sufficiently well calibrated to drive automated experiment-planning—removing a key barrier to closed-loop, self-optimising test campaigns. These findings suggest that combining machine learning models with active learning strategies is a powerful approach to building faster, more responsive and efficient DT. By optimizing both model structure and data acquisition, DTs can achieve faster adaptation, improved accuracy, and better real-time decision-making support — critical capabilities for next-generation nuclear, energy, and industrial systems.

*6.2. Sensor assimilation and Spatial Generalisation*

High-temperature reactor facilities such as the OSU-HTTF contain hundreds of thermocouples distributed irregularly in space. Engineering judgment typically dictates sensor placement, yet outages and budgetary constraints inevitably lead to partial coverage. Our structured concatenation algorithm reorganizes disparate sensor streams into a single quasi-periodic sequence, which the variational GRU (VGRU) then learns with minimal pre-processing. Even when 50 % of the sensors were withheld during training, the VGRU reproduced both solid- and fluid-temperature fields with $R^2$ >0.99 and delivered smooth, credible uncertainty bounds. The sharper degradation observed for the more parameter-rich variational LSTM (VLSTM) highlights a practical advantage of GRU cells: their simpler gating makes them less sensitive to dimensionality collapse and therefore better suited for assets where instrumentation may degrade over time. For plant owners, this implies a path to "graceful aging" of digital twins—maintaining accurate state estimation, and hence situational awareness, even as sensor networks thin or re-configure.

*6.3. Iterative Updating for Dynamic Digital Twins*

Renewable-energy-generation fleets and electrochemical storage systems evolve continuously: seasonal irradiance patterns drift, turbines experience component upgrades, and batteries suffer progressive capacity fade. Static models trained once on historical archives quickly become obsolete. The PSML and Li-ion studies show that a VDT equipped with rolling variational updates can track such drifts with modest compute budgets. In the PSML benchmark, both VGRU and VLSTM retained higher performance for solar power under a monthly update cadence, while wind errors—which are dominated by rare regime shifts—grew only slowly, confirming the need for longer context windows in that channel. The battery twin, based on a physics-informed BattNN backbone, offers perhaps the most compelling evidence: mean-squared error dropped by an order of magnitude relative to the best static alternative as end-of-life behavior emerged. For operators, such well-calibrated predictive uncertainty supports probabilistic dispatch decisions, dynamic reserve allocation, and condition-based maintenance scheduling.

*6.4. Cross-cutting Challenges and Implications*

Several pragmatic hurdles commonly cited in digital-twin deployments were addressed. First, *computational tractability*: variational last layers add only two trainable parameters per output unit, and our GPU timings (tens of seconds per update) confirm their suitability for near-edge inference. Second, *data-scarce calibration*: the CHF and HTTF results show that the same Bayesian machinery underpins both greedy experiment selection and robust learning from sparse sensor fields. Third, *trust and interpretability*: by providing credible intervals alongside every forecast, the VDT offers a transparent risk metric that can be embedded in safety cases or operational limits—an essential requirement in highly regulated sectors such as nuclear and grid management.

For academic investigators, the modularity demonstrated here invites systematic studies on (i) optimal retraining cadence versus process time-scale, (ii) hybrid schemes that update both variational layers and embedded physical parameters, and (iii) domain-specific priors that further tighten uncertainty bounds. Industrial practitioners, on the other hand, may see immediate value in three deployment patterns: (1) AAL-driven test campaigns that cut qualification time for new fuel or materials; (2) reactor and process twins that continue to function amid partial sensor loss; and (3) asset-health twins for batteries and renewables that supply probabilistic forecasts compatible with existing energy-management systems. Collectively, the case studies argue that lightweight variational Bayesian augmentation—rather than wholesale model redesign—can be sufficient to elevate conventional machine-learning surrogates into fully fledged, uncertainty-aware digital twins capable of safe, long-horizon operation in complex energy infrastructures.

## 7. Conclusions

This work proposed a modular, uncertainty-aware variational digital-twin framework that couples lightweight variational inference with task-specific neural backbones. The inverse loop—responsible for real-time data assimilation was implemented by attaching a Bayesian last layer to otherwise standard feed-forward, recurrent, and physics-informed networks. In four energy-sector case studies, this simple yet principled modification delivered (i) calibrated predictive intervals at negligible computational cost, (ii) rapid active-learning gains when data are scarce, (iii) robust state estimation from sparse or degraded sensor fields, and (iv) stable long-horizon forecasting under non-stationary operating conditions.

In the critical-heat-flux problem, VFNN uncertainty estimates guided an aided active-learning strategy that reached the regulatory accuracy threshold with fewer experiments and one-third the training time required by random sampling. The HTTF reactor study showed that a structured concatenation algorithm combined with a VGRU could assimilate hundreds of heterogeneous temperature streams, retain high predictive accuracy after 50% sensor attrition, and produce interpretable confidence bands—capabilities essential for safety-critical monitoring. For renewable energy generation, monthly retraining of variational RNNs maintained for solar power over three years while containing the natural error growth seen in highly volatile wind signals. Finally, the Li-ion battery twin, built on a physics-informed BattNN backbone, reduced mean-squared error by an order of magnitude relative to the best static alternative and supplied credible intervals that adapted smoothly from early life to end of life.

These results resolve several practical barriers to digital-twin deployment noted in the literature review: the VDT achieves real-time calibrations on commodity GPUs; converts predictive variance into a utility measure for data acquisition; and remains functional under instrumentation loss and regime shifts. For academic researchers, the study offers a reproducible template for integrating variational layers into pre-existing models and a set of benchmarks that isolate active learning, sensor assimilation, and life-cycle updating. For industrial practitioners, it provides quantitative evidence that modest Bayesian augmentation can turn conventional surrogates into trustworthy, online twins suitable for grid operations, process supervision, and asset-health management.

Several avenues merit further investigation. First, integrating periodic re-identification of embedded physical parameters—especially in physics-informed networks—could reduce or eliminate the residual error growth observed in very long deployments. Second, the generative loop outlined in Section [2](#) will be developed to reconstruct full-field states from sparse measurements, enhancing spatial resolution without additional sensors. Third, rigorous assessment of prior choices, retraining cadence, and hierarchical variational structures across domains will clarify best practices for uncertainty calibration. Finally, closed-loop studies that couple the inverse VDT with a decision-making forward controller will extend the framework from passive monitoring to autonomous control. The variational digital-twin concept presented here offers a computationally tractable, statistically principled pathway toward real-time, self-updating models of complex energy assets. By unifying Bayesian learning, active data acquisition, and scalable neural architectures, the framework bridges the gap between academic prototypes and industrially robust cyber-physical systems. With AI now driving unprecedented energy consumption, the time has come to deploy its strengths in service of smarter, more efficient energy infrastructures.

## Data Availability

The authors have all the data and codes to reproduce all the results in this work currently in a private GitHub repository. To ensure the confidentiality of this research, the authors will make this repository public during an advanced stage of the review process, which will be listed under our research group's public Github page: <https://github.com/aims-umich>

## Acknowledgment

This work received sponsorship through the National Science Foundation's Graduate Research Fellowship Program (Grant Number: DGE 2241144). This work was also supported through Idaho National Laboratory's Laboratory Directed Research and Development (LDRD) Program (Award Number: 24A1081-116FP) under Department of Energy Idaho Operations Office contract no. DE-AC07-05ID14517.

## CRediT Author Statement

- **Logan A. Burnett**: Conceptualization, Methodology, Software, Validation, Formal Analysis, Visualization, Investigation, Data Curation, Writing - Original Draft.

- **Umme Mahbuba Nabila**: Methodology, Software, Data Curation, Formal Analysis, Visualization, Investigation, Writing - Original Draft.

- **Majdi I. Radaideh**: Conceptualization, Methodology, Resources, Funding Acquisition, Supervision, Project Administration, Writing - Original Draft.